\documentclass[10pt,twocolumn,letterpaper]{article}

\usepackage{wacv}
\usepackage{times}
\usepackage{epsfig}
\usepackage{graphicx}
\usepackage{amsmath}
\usepackage{amssymb}
\usepackage{booktabs}
\usepackage{rotating}
\usepackage{makecell}
\usepackage{xfrac}
\usepackage[dvipsnames]{xcolor}
\usepackage{amsthm}
\usepackage{comment}
\usepackage{nicefrac}
\usepackage{xargs}
\usepackage{xspace}

\usepackage[accsupp]{axessibility} 

\newcommand{\aka}{\emph{a.k.a.}\xspace}

\newtheorem*{example*}{Example}

\newtheorem*{examples*}{Examples}

\definecolor{myblue}{HTML}{1f77b4}
\definecolor{myorange}{HTML}{ff7f0e}
\definecolor{myred}{HTML}{d62728}
\definecolor{mypurple}{HTML}{9467bd}
\definecolor{mygreen}{HTML}{2ca02c}

\definecolor{mybluebis}{HTML}{57c2e6} 
\definecolor{myorangebis}{HTML}{ffbf00}

\def\wacvPaperID{11} 

\wacvapplicationstrack 

\wacvfinalcopy 

\ifwacvfinal
\usepackage[breaklinks=true,bookmarks=false]{hyperref}
\else
\usepackage[pagebackref=true,breaklinks=true,colorlinks,bookmarks=false]{hyperref}
\fi

\begin{document}

\newcommand{\myblue}[1]{\textbf{\textcolor{myblue}{#1}}}
\newcommand{\myorange}[1]{\textbf{\textcolor{myorange}{#1}}}
\newcommand{\myred}[1]{\textbf{\textcolor{myred}{#1}}}
\newcommand{\mypurple}[1]{\textbf{\textcolor{mypurple}{#1}}}
\newcommand{\mygreen}[1]{\textbf{\textcolor{mygreen}{#1}}}

\newcommand{\mybluebis}[1]{\textbf{\textcolor{mybluebis}{#1}}}
\newcommand{\myorangebis}[1]{\textbf{\textcolor{myorangebis}{#1}}}

\global\long\def\sampleSpace{\Omega}%
\global\long\def\aSample{s}%
\global\long\def\eventSpace{\Sigma}%
\global\long\def\anEvent{X}%

\global\long\def\entropy{H}

\global\long\def\weightUnconditional{\lambda}%
\global\long\def\vectorWeightUnconditional{\mathbf{\lambda}}%
\global\long\def\weightConditional{\omega}%
\global\long\def\vectorWeightConditional{\mathbf{\omega}}%
\global\long\def\arbitrarilyChosenWeightUnconditional{\kappa}%
\global\long\def\vectorArbitrarilyChosenWeightUnconditional{\mathbf{\kappa}}%

\global\long\def\allEvidences{\mathbb{E}}%
\global\long\def\anEvidence{E}%
\global\long\def\allHypotheses{\mathbb{H}}%
\global\long\def\anHypothesis{H}%

\global\long\def\allDomainsSource{\mathbb{D}_S}%
\global\long\def\allDomainsTarget{\mathbb{D}_T}%
\global\long\def\domain{d}%
\global\long\def\domainSource{{\domain_S}}%
\global\long\def\domainTarget{{\domain_T}}%
\global\long\def\numDomainSource{n}%

\global\long\def\probabilityMeasure{P}%
\global\long\def\overallProbabilityMeasure{P}%
\global\long\def\realNumbers{\mathbb{R}}%
\global\long\def\allPlausibleSources{\Phi}%
\newcommandx\eventDomainSource[1][usedefault, addprefix=\global, 1=\idxDomainSource]{D_{S_{#1}}}%
\global\long\def\idxDomainSource{k}%
\global\long\def\simplexSemanticClasses{{\mathbf{\Delta}^{\left|\allSemanticClasses\right|-1}}}%
\global\long\def\simplexDomainSource{{\mathbf{\Delta}^{\numDomainSource-1}}}%
\global\long\def\interior#1{\mathring{#1}}%
\global\long\def\simplexDomainSourceInterior{{\interior{\mathbf{\Delta}}^{\numDomainSource-1}}}%


\global\long\def\performanceSampleSpace{\Omega_{\mathcal{P}}}%
\global\long\def\performanceEventSpace{\Sigma_{\mathcal{P}}}%
\global\long\def\allPerformanceMeasures{\mathbb{P}}%
\global\long\def\randVarSatisfaction{S}%

\global\long\def\randVarGroundtruthClass{C}%
\global\long\def\randVarPredictedClass{\hat{C}}%
\global\long\def\allClasses{\mathbb{C}}%
\global\long\def\aClass{c}%

\global\long\def\performanceProbabilityMeasureUnbalanced{\mathcal{P}}%
\global\long\def\performanceProbabilityMeasureBalanced{\mathcal{P^*}}%

\global\long\def\indicatorWeightedAccuracy{WA}%
\global\long\def\indicatorAccuracy{A}%
\global\long\def\indicatorTrueRate#1{TR_{#1}}%
\global\long\def\indicatorJaccard#1{J_{#1}}%
\global\long\def\indicatorBalancedAccuracy{BA}%
\global\long\def\indicatorMeanIOU{mIoU}%
\global\long\def\indicatorBalancedMeanIOU{BmIoU}%
\global\long\def\indicatorAny{I}%
\global\long\def\indicatorBalancedAny{BI}%

\global\long\def\balance{\mathrm{balance}}%

\newcommand{\mysection}[1]{\vspace{2pt}\noindent\textbf{#1}}

\global\long\def\allImages{\mathbb{I}}%
\global\long\def\anImage{i}%
\global\long\def\allPixels{\mathbb{P}}%
\global\long\def\aPixel{p}%
\global\long\def\allSemanticClasses{\mathbb{S}}%
\global\long\def\aSemanticClass{s}%

\global\long\def\importanceOfGroundtruthSemanticClass{\mathcal{I}}%

\global\long\def\groundtruthSemanticClass{Y}%
\global\long\def\predictedSemanticClass{\hat{Y}}%

\global\long\def\noiseSourceModels{\epsilon_\domainSource}%
\global\long\def\noiseDomainDiscriminatorModel{\epsilon_{\vectorWeightConditional_{\anEvidence}}}%
\global\long\def\noiseOutput{\epsilon_\domainTarget}%

\global\long\def\functionExactModel{f_{\anHypothesis}}%
\global\long\def\functionExactModelUpToTargetShift{f^{*}_{\anHypothesis}}%
\global\long\def\functionSourceModel{f^{*}_{sm}}%
\global\long\def\functionDomainDiscriminatorModel{f^{*}_{ddm}}%

\global\long\def\dot{\,.}%
\global\long\def\comma{\,,}%


\title{Mixture Domain Adaptation to Improve\\
Semantic Segmentation in Real-World Surveillance}

\author{Sébastien Piérard\\
University of Liège, Belgium\\
{\tt\small S.Pierard@uliege.be}
\and
Anthony Cioppa\\
University of Liège, Belgium\\
{\tt\small Anthony.Cioppa@uliege.be}
\and
Anaïs Halin\\
University of Liège, Belgium\\
{\tt\small Anais.Halin@uliege.be}
\and
Renaud Vandeghen\\
University of Liège, Belgium\\
{\tt\small R.Vandeghen@uliege.be}
\and
Maxime Zanella\\
University of Louvain-la-Neuve, Belgium\\
{\tt\small Maxime.Zanella@uclouvain.be}
\and
Benoît Macq\\
University of Louvain-la-Neuve, Belgium\\
{\tt\small Benoit.Macq@uclouvain.be}
\and
Saïd Mahmoudi\\
University of Mons, Belgium\\
{\tt\small Said.Mahmoudi@umons.ac.be}
\and
Marc Van Droogenbroeck\\
University of Liège, Belgium\\
{\tt\small M.VanDroogenbroeck@uliege.be}
}

\maketitle
\thispagestyle{empty}

\begin{abstract}
    Various tasks encountered in real-world surveillance can be addressed by determining posteriors (e.g. by Bayesian inference or machine learning), based on which critical decisions must be taken. However, the surveillance domain (acquisition device, operating conditions, etc.) is often unknown, which prevents any possibility of scene-specific optimization. In this paper, we define a probabilistic framework and present a formal proof of an algorithm for the unsupervised many-to-infinity domain adaptation of posteriors. Our proposed algorithm is applicable when the probability measure associated with the target domain is a convex combination of the probability measures of the source domains. It makes use of source models and a domain discriminator model trained off-line to compute posteriors adapted on the fly to the target domain. Finally, we show the effectiveness of our algorithm for the task of semantic segmentation in real-world surveillance. The code is publicly available at \url{https://github.com/rvandeghen/MDA}.
\end{abstract}


\section{Introduction}

\begin{figure}
    \centering
    \includegraphics[width=0.9\linewidth]{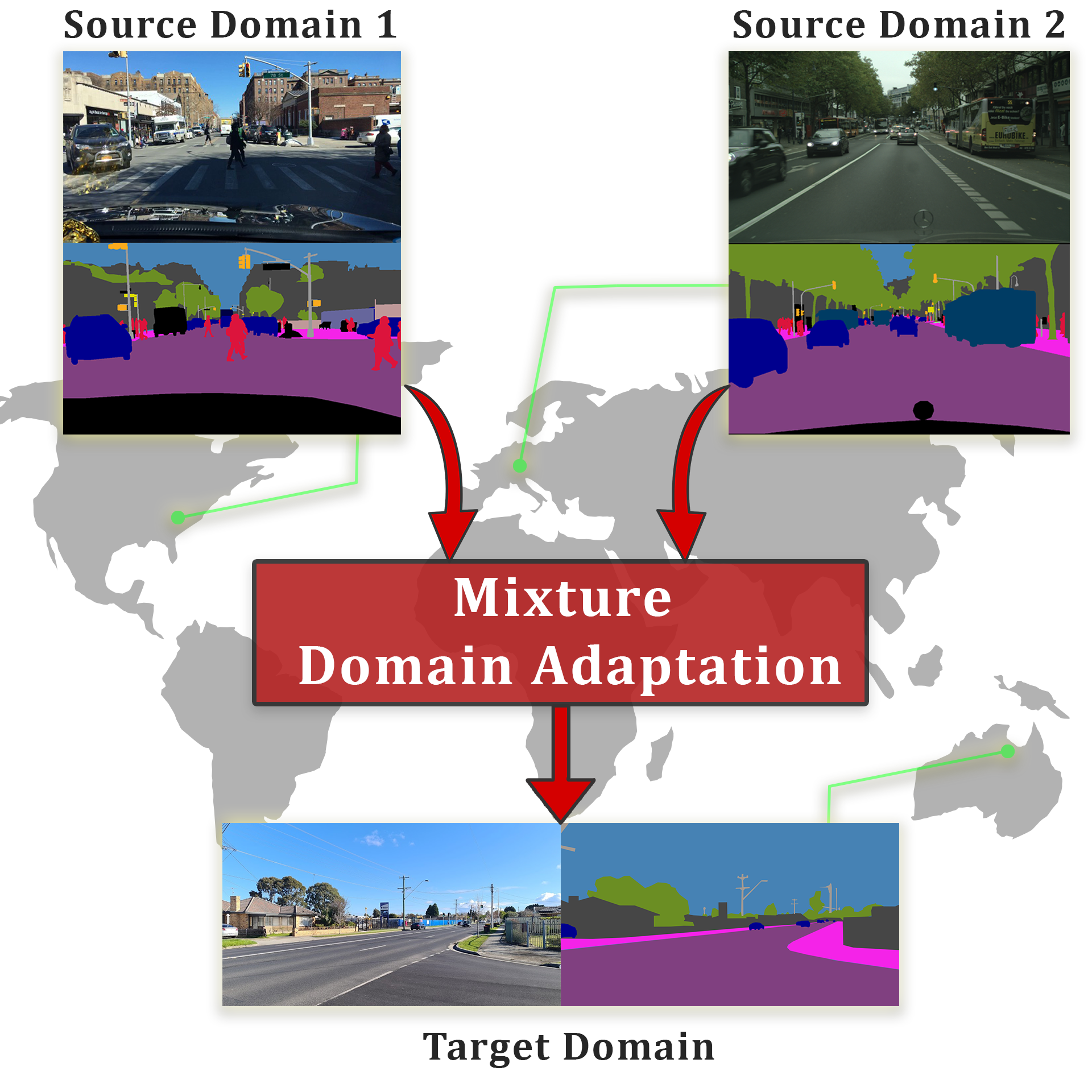}
    \caption{We present a theoretically motivated algorithm for unsupervised domain adaptation of posteriors on the fly. 
    In particular, we express the target domain as a convex combination of the source domains. 
    We show the effectiveness of our algorithm for the task of semantic segmentation in real-world surveillance.} 
    \label{fig:my_label}
\end{figure}

Applying artificial intelligence methods to real-world surveillance requires taking into account the specificity of the considered scene. In this paper, we address this issue by computing posteriors, which are probabilities related to the content of the scene. For example, for the task of semantic segmentation, the posteriors are the conditional probabilities of the semantic classes (the \emph{hypotheses}) given a pixel from an image (an \emph{evidence}).

There are three main reasons for focusing on posteriors: (1) they help interpret results compared to hard decisions (\eg classifying an object with some degree of confidence rather than simply claiming it belongs to a class), (2) they enable optimal decisions for most common criteria (\eg maximizing the posteriors for the accuracy or maximizing the likelihoods for the balanced accuracy), and (3) they allow to adapt to the specific domain of a scene, as demonstrated in this paper. 

In general, the domains are rarely known in advance, which prevents any possibility of scene-specific optimization. Indeed, the domains depend on the choices made for the data acquisition (\eg where the camera is installed or when the videos or images are captured) as well as on the distribution of scenarios that can occur (\eg varying weather or lighting conditions). 
Furthermore, in some applications, the domains may change over time, making it even more challenging to take into account the specificity of the scene. This typically occurs when the surveillance cameras are onboard of moving vehicles.

When dealing with multiple domains, a training dataset can be collected in each domain. Common but oversimplistic assumptions are that the testing distribution (referred to as the \emph{target domain}) is a fixed mixture of the various training datasets (referred to as the \emph{source domains}) and that the mixture parameters are known in advance. Under these assumptions, it is reasonable to mix the various training datasets before training the models. This, however, is impossible in most real-world cases as the mixture parameters depend on the target domain which is unknown at learning time. An on-the-fly and unsupervised domain adaptation is therefore required for the sake of flexibility.

In this paper, we propose an algorithm that makes use of \emph{source models} and a \emph{domain discriminator model} trained off-line to obtain posteriors adapted on the fly to the target domain. We present a formal proof of this algorithm that is interpretable and runs in real time. Its effectiveness is shown experimentally for the task of semantic segmentation in real-world surveillance.

\mysection{Contributions.} Our contributions are as follows.
 \textbf{(i)} We define a probabilistic framework for a particular unsupervised mixture domain adaptation problem.
\textbf{(ii)} Based on this framework, we present a formal proof of an unsupervised algorithm to adapt the posteriors on the fly for a changing target domain.
\textbf{(iii)} We compare our algorithm with common posterior combination heuristics and show its superiority on several real-world surveillance datasets.

\section{Related work\label{sec:related-work}}

Hereafter, we present the related work first on unsupervised domain adaptation and second on semantic segmentation, with some references specific to the field of real-world surveillance.

\mysection{Unsupervised domain adaptation.} Domain adaptation aims at transferring knowledge from source domains to unseen  target domains and alleviating the impact of domain shift in data distributions~\cite{Farahani2021ABrief}. An even more challenging scenario arises when no label is available in the target domains; this is called unsupervised domain adaptation~\cite{Liu2022Deep}.
Domain shifts, which are defined as a change in the data distribution between the training distribution of an algorithm and the inference distribution, can be categorized into three main categories~\cite{Farahani2021ABrief}: (1) \textit{prior shift} (\aka \emph{target shift}), when the priors are different but the likelihoods are identical~\cite{Sipka2022TheHitchhikerGuide}; (2) \textit{covariate shift}, when the marginal probability distributions differ between the source and target domains, but not the posteriors~\cite{Kirchmeyer2022Mapping}; and (3) \textit{concept shift}, when data distributions remain unchanged across source and target domains, while the posteriors differ between domains~\cite{Redko2019Advances}.

A common solution to unsupervised domain adaptation consists in adding a domain discriminator to learn domain-independent features~\cite{Ganin2015Unsupervised}. This approach needs a new training phase, making real-time predictions unfeasible. Moreover, aligning the features can reduce their discriminative power~\cite{Saito2018Maximum}. A more restrictive scenario can be investigated, in which the source domain data cannot be accessed at test-time. Liang \etal~\cite{Liang2020DoWe} freeze the classifier and learn representations from the target domains aligned to the source hypothesis. Wang \etal~\cite{Wang2020Tent} adapt the normalization layers while minimizing the prediction entropy. Boudiaf \etal~\cite{Boudiaf2022Parameterfree} propose to keep the parameters of the model unchanged to avoid collapse in case of poorly diversified batches and only correct the predictions by encouraging neighboring representations to have consistent predictions.  

Other works tackle the problem of domain shifts in the case of semantic segmentation. Zhang \etal~\cite{Zhang2017Curriculum} suggest to use a curriculum learning approach to learn more transferable global and local properties across domains. Li \etal~\cite{Li2019Bidirectional} propose a bidirectional learning framework in which a translation module and a segmentation model are trained alternatively.


In the context of multi-source semantic segmentation, unsupervised domain adaptation has been studied with some approaches generating adapted domains~\cite{Zhao2019MultiSource,Zhao2021MADAN} or aligning their features distribution by matching their moments~\cite{Peng2019Moment}. He \etal~\cite{He2021MultiSource} propose to reduce the discrepancy between domains by aligning their pixel distribution before training models in a collaborative way to share knowledge between sources. One of the closest works to ours for binary classification is presented in~\cite{Sun2013Bayesian}. The authors derive posteriors from Bayes' theorem by approximating priors with distances to each point in each source domain and likelihoods from nearest neighbor points. In our work, we address a multi-class case in a dense task, \ie semantic segmentation. We further characterize the target domain as a mixture of the source domains.

\mysection{Semantic segmentation.} For the past decade, semantic segmentation has proven to be a powerful tool for global scene understanding~\cite{GarciaGarcia2017AReview,Liu2018Recent,Minaee2021Image}. 
To support the development of semantic segmentation networks, many annotated datasets emerged such as ADE20K~\cite{Zhou2017Scene, Zhou2018Semantic}, PascalVOC2012~\cite{Everingham2011PASCAL}, COCO~\cite{Lin2014COCO}, or CityScapes~\cite{Cordts2016The}. Their availability has boosted the performance of algorithms over the years, leading to algorithms such as PSPNet~\cite{Zhao2017Pyramid}, Mask R-CNN~\cite{He2017MaskRCNN}, PointRend~\cite{Kirillov2020PointRend}, or SETR~\cite{Zheng2021Rethinking}. 

In the field of real-world surveillance, semantic segmentation can form the basis of complex downstream tasks such as urban scene characterization~\cite{Zhou2021GMNet}, maritime surveillance~\cite{Cane2018Evaluating}, water level estimation~\cite{Muhadi2021Deep}, low-light video enhancement~\cite{Zheng2022SemanticGuided}, or super-resolution~\cite{Aakerberg2022Semantic}. Due to the variety of scenes and downstream tasks, it is difficult to develop semantic segmentation algorithms that remain competitive without further scene-specific tuning.
%
For instance, when placing a new surveillance camera, we need to ensure that the segmentation network will be able to perform well in this novel environment. 
Furthermore, the network must be robust to dynamic domain changes. Short-term changes may include illumination changes, sudden heavy rains, or different occupancy during rush hours, while long-term changes may include day-lasting heavy snow during winter or road constructions that last for months.
If the network has not been trained on all those particular domains, it may simply fail to predict the correct labels and therefore lead to critical failures.
Unfortunately, the networks that are robust to a wide variety of domains are often too large, may not fit in memory, may not be fast enough for real-time processing, and may require high power consumption.

One naive scene-specific adaptation consists in training a small real-time network on scene-specific data in a supervised fashion.
However, this requires collecting a dataset prior to installation, which may be impractical.
Also, if the domain changes dynamically, the network may be unable to adapt.
As a solution, Cioppa~\etal~\cite{Cioppa2019ARTHuS} propose the online distillation framework, in which a large network is used to train a small real-time network on the fly. 
This allows the real-time network to adapt to dynamically changing domains without requiring any manual annotation.
We propose an alternative for dynamic domain adaptation by leveraging real-time networks trained on various domains.

\section{Method\label{sec:method}}

\subsection{Probabilistic framework}
\label{subsec:probabilistic-framework}

\mysection{Similarities and differences between domains.}
By assumption, all domains share a measurable space $\left(\sampleSpace,\eventSpace\right)$, where $\sampleSpace$ is a non-empty set (the \emph{sample space}, or \emph{universe}) and $\eventSpace\subseteq\sampleSpace^{2}$ is a $\sigma$-algebra on it (the \emph{event space}). They also share a non-empty set of \emph{evidences}, $\allEvidences\subseteq\eventSpace$, and a non-empty set of \emph{hypotheses}, $\allHypotheses\subseteq\eventSpace$. The differences between the various domains stand in the probability measures. We denote the one associated with the domain $\domain$ by $\probabilityMeasure_{\domain}$.%

\mysection{Bayesian inference.}
This task consists in determining the posterior $\probabilityMeasure\left(\anHypothesis\mid\anEvidence\right)$ for any given hypothesis $\anHypothesis\in\allHypotheses$ and any given evidence $\anEvidence\in\allEvidences$.  In the following, entities performing this task are called \emph{models}. In some cases, \emph{exact models} $\functionExactModel:\allEvidences\rightarrow\realNumbers:\anEvidence\mapsto\probabilityMeasure\left(\anHypothesis\mid\anEvidence\right)$ can be established theoretically.
In Section~\ref{sec:method}, we relax this constraint and assume that the models are \emph{exact up to a target shift}, \ie $\functionExactModelUpToTargetShift:\allEvidences\rightarrow\realNumbers:\anEvidence\mapsto\probabilityMeasure^{*}\left(\anHypothesis\mid\anEvidence\right)$, 
 the probability measures $\probabilityMeasure$ and $\probabilityMeasure^{*}$ having equal likelihoods:
\begin{equation}
    \probabilityMeasure\left(\anEvidence\mid\anHypothesis\right)
    =
    \probabilityMeasure^*\left(\anEvidence\mid\anHypothesis\right)
    \qquad
    \forall \anEvidence \in \allEvidences,\,
    \forall \anHypothesis \in \allHypotheses
    \dot
    \label{eq:target-shift-hypothesis}
\end{equation}
In case of non-zero priors, one can recover an exact model by applying a correction, called \emph{target shift}, to the output $\probabilityMeasure^{*}\left(\anHypothesis\mid\anEvidence\right)$. When $\allHypotheses$ forms a partition of $\sampleSpace$, this correction~\cite{Sipka2022TheHitchhikerGuide,Pierard2014OnTheFly} can be written as
\begin{equation}
    \probabilityMeasure\left(\anHypothesis\mid\anEvidence\right)
    =
    \frac{
        \frac
        {\probabilityMeasure\left(\anHypothesis\right)}
        {\probabilityMeasure^*\left(\anHypothesis\right)}
        \,
        \probabilityMeasure^*\left(\anHypothesis\mid\anEvidence\right)
    }{
        \sum_{\anHypothesis\in\allHypotheses}
        \frac
        {\probabilityMeasure\left(\anHypothesis\right)}
        {\probabilityMeasure^*\left(\anHypothesis\right)}
        \,
        \probabilityMeasure^*\left(\anHypothesis\mid\anEvidence\right)
    }
    \dot
    \label{eq:target-shift}
\end{equation}

\mysection{Geometric representation.}
When $\allHypotheses$ is finite and forms a partition of $\sampleSpace$, priors and posteriors can be represented by points in any non-degenerated Euclidean simplex in $\left|\allHypotheses\right|-1$ dimensions (\eg a point, a segment, a triangle, a tetrahedron). After establishing a bijection between its vertices and $\allHypotheses$, the probability relative to the hypothesis $\anHypothesis$ (either the prior or the posterior) can be obtained by projecting the point on the axis that passes through the vertex $\anHypothesis$ and that is orthogonal to its opposite face, the probability being equal to $0$ on the face and to $1$ at the vertex.

\begin{figure}
\begin{centering}
\newcommand{\textA}{$\anHypothesis_1$}
\newcommand{\textB}{$\anHypothesis_2$}
\newcommand{\textC}{$\anHypothesis_3$}
\newcommand{\textD}{maximum a posteriori (MAP)}
\newcommand{\textE}{maximum likelihood estimation (MLE)}
\resizebox{1\linewidth}{!}{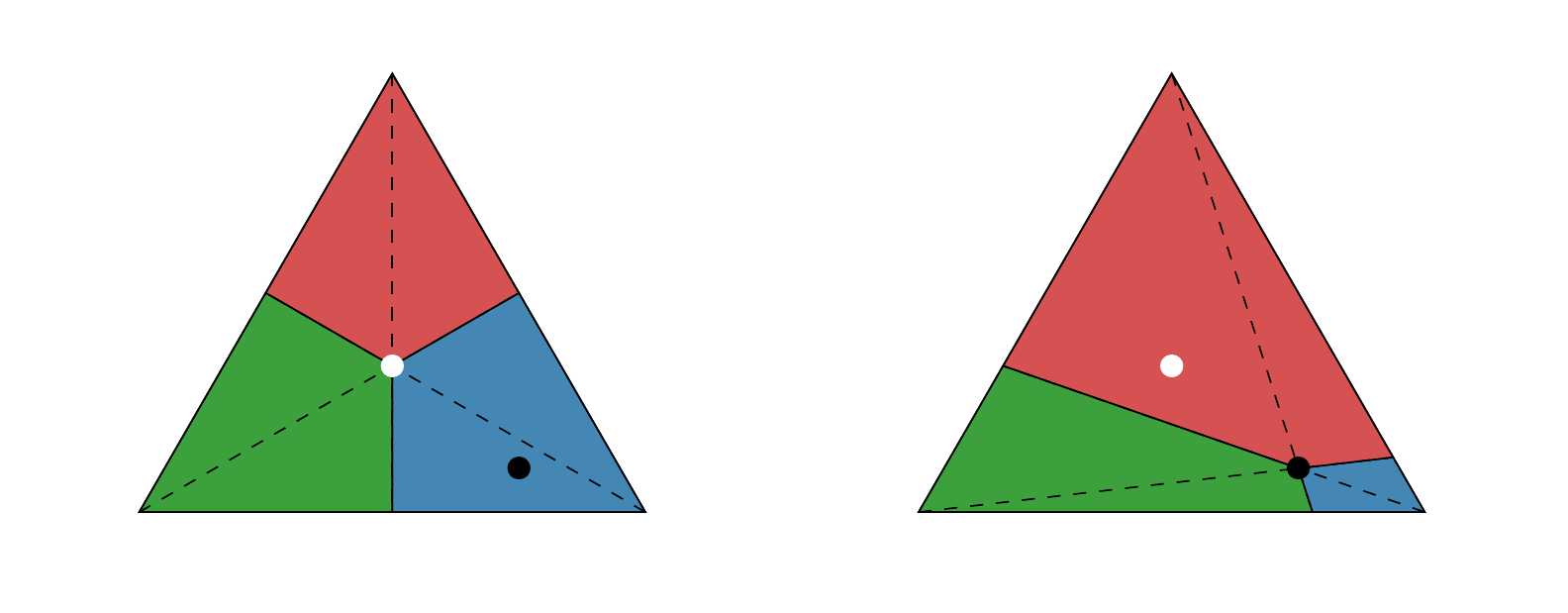}
\end{centering}
\caption{Comparison between two decision-making strategies: maximizing the accuracy (MAP, left) and maximizing the balanced accuracy (MLE, right). Here, the three colors represent the decisions for the three hypotheses $\allHypotheses=\left\{\anHypothesis_1,\anHypothesis_2,\anHypothesis_3\right\}$ forming a partition of $\sampleSpace$. The black dot represents the priors, arbitrarily taken as  $\probabilityMeasure\left(\anHypothesis_1\right)=0.1$, $\probabilityMeasure\left(\anHypothesis_2\right)=0.2$, and $\probabilityMeasure\left(\anHypothesis_3\right)=0.7$.}\label{fig:decision_making_in_triangle}
\end{figure}

\mysection{Decision-making.}
In any domain $\domain$, making a decision for a given evidence $\anEvidence$ comes to choose a hypothesis based on the posteriors $\probabilityMeasure_\domain\left(\anHypothesis\mid\anEvidence\right)\,\forall\anHypothesis\in\allHypotheses$. Together, the Bayesian inference and the decision-making form a function $\allEvidences\rightarrow\allHypotheses$. We recall two standard decision-making strategies.
\begin{itemize}
    \item MAP (\emph{maximum a posteriori}) selects the hypothesis with the highest posterior. MAP has a discontinuity where the hypotheses are equally likely given the evidence. It maximizes the probability of making a correct decision, \ie the accuracy.
    \item MLE (\emph{maximum likelihood estimation}) selects the hypothesis with the highest likelihood (or, equivalently, the highest posterior to prior ratio). MLE has a discontinuity where posteriors and priors are equal, \ie when the evidence provides no information about the hypotheses. It maximizes the balanced accuracy.
\end{itemize}
As shown in Figure~\ref{fig:decision_making_in_triangle}, both strategies partition the Euclidean simplex into $\left|\allHypotheses\right|$ convex parts. These two strategies are related to each others by the fact that the balanced accuracy corresponds to the accuracy after shifting the priors to equally likely hypotheses.

\subsection{Problem statement}
\label{subsec:problem-statement}

Let us now focus on real-world problems in which there are two kinds of domains: the \emph{source} and \emph{target} domains, given respectively by the sets $\allDomainsSource$ and $\allDomainsTarget$. In all source domains $\domainSource\in\allDomainsSource$, we assume that it is possible to obtain models computing or estimating the posteriors $\probabilityMeasure_{\domainSource}\left(\anHypothesis\mid\anEvidence\right)$, possibly after gathering annotated data following the distribution $\probabilityMeasure_{\domainSource}\left(\anEvidence,\anHypothesis\right)$, to the contrary of any ``new" target domain $\domainTarget \in \allDomainsTarget \setminus \allDomainsSource$. The computation of $\probabilityMeasure_{\domainTarget}\left(\anHypothesis\mid\anEvidence\right)$ is an \emph{unsupervised domain adaptation} problem.


We consider now the particular many-to-infinity domain adaptation problem in which the probability measure of any target domain $\domainTarget\in\allDomainsTarget$ can be obtained as a convex combination of the probability measures of the source domains in the non-empty set $\allDomainsSource=\left\{\domainSource_1, \domainSource_2, \ldots \domainSource_\numDomainSource\right\}$ as follows
\begin{equation}
    \probabilityMeasure_{\domainTarget}
    =
    \sum_{\idxDomainSource=1}^{\numDomainSource}
    \weightUnconditional_{\idxDomainSource}
    \probabilityMeasure_{\domainSource_{\idxDomainSource}}
    \comma
    \label{eq:domain-target-fct-sources}
\end{equation}
with $\vectorWeightUnconditional=(\weightUnconditional_1,\weightUnconditional_2\ldots \weightUnconditional_{\numDomainSource})\in\simplexDomainSource$.\footnote{We use the notation $\simplexDomainSource$ to denote the $(\numDomainSource-1)$-probabilistic simplex, that is $\vectorWeightUnconditional\in\simplexDomainSource$ if and only if $\sum_{\idxDomainSource=1}^{\numDomainSource}\weightUnconditional_\idxDomainSource=1$ and $\weightUnconditional_{\idxDomainSource}\ge0, \, \forall \idxDomainSource$.} This problem is complex as the distribution of evidences $\probabilityMeasure\left(\anEvidence\right)$, the priors $\probabilityMeasure\left(\anHypothesis\right)$, and the likelihoods (\aka appearance models) $\probabilityMeasure\left(\anEvidence \mid \anHypothesis\right)$ may vary from one domain to another.

From Equation~\ref{eq:domain-target-fct-sources}, it results that the distribution of evidences in the target domain is a mixture of the distributions of evidences in the source domains:
$
    \probabilityMeasure_{\domainTarget}\left(\anEvidence\right)
    =
    \sum_{\idxDomainSource=1}^{\numDomainSource}
    \weightUnconditional_{\idxDomainSource}
    \probabilityMeasure_{\domainSource_{\idxDomainSource}}\left(\anEvidence\right)
$, $\forall \anEvidence \in \allEvidences$. Therefore, the domain adaptation problem studied in this paper is a particular case of the more general \emph{mixture adaptation problem} introduced by Mansour~\etal in~\cite{Mansour2008Domain}.

By construction, we consider that the priors can be pre-computed for source domains. Consequently, the priors in the target domain can be computed by $
    \probabilityMeasure_{\domainTarget}\left(\anHypothesis\right)
    =
    \sum_{\idxDomainSource=1}^{\numDomainSource}
    \weightUnconditional_{\idxDomainSource}
    \probabilityMeasure_{\domainSource_{\idxDomainSource}}\left(\anHypothesis\right)
$, $\forall \anHypothesis \in \allHypotheses$. Moreover, the likelihoods in the target domain are mixtures of the corresponding likelihoods from the source domains as $
    \probabilityMeasure_{\domainTarget}\left(\anEvidence\mid\anHypothesis\right)
    =
    \sum_{\idxDomainSource=1}^{\numDomainSource}
    \weightConditional_{\idxDomainSource,\anHypothesis}
    \probabilityMeasure_{\domainSource_{\idxDomainSource}}\left(\anEvidence\mid\anHypothesis\right)
$ with $\weightConditional_{\idxDomainSource,\anHypothesis}=\weightUnconditional_{\idxDomainSource}\nicefrac{\probabilityMeasure_{\domainSource_{\idxDomainSource}}\left(\anHypothesis\right)}{\probabilityMeasure_{\domainTarget}\left(\anHypothesis\right)}$, $\forall \anEvidence \in \allEvidences$ and $\forall \anHypothesis \in \allHypotheses$. Therefore, the domain adaptation problem studied here is a sub-case of a more general one in which the target likelihoods are  mixtures of fixed components (in our case, the likelihoods in the source domains), the component weights and the target priors being known only at runtime. Some theoretical and experimental results for that on-the-fly domain adaptation problem can be found in \cite{Pierard2014OnTheFly}, in the very specific case of two-class classifiers.

\subsection{Problem analysis and theoretical solution}
\label{subsec:problem-analysis-and-solution}

Let us consider any evidence $\anEvidence$ such that $\probabilityMeasure_{\domainTarget}\left(\anEvidence\right) \ne 0$. 
From Equation~\ref{eq:domain-target-fct-sources} and using probability theory, we can derive an exact formula to compute $\probabilityMeasure_{\domainTarget}\left(\anHypothesis \mid \anEvidence\right)$ in any target domain $\domainTarget\in\allDomainsTarget$. We have, $\forall \anHypothesis \in \allHypotheses$,
\begin{equation}
    \probabilityMeasure_{\domainTarget}\left(\anHypothesis \mid \anEvidence\right)
    =
    \sum_{\domainSource_{\idxDomainSource}\in\allPlausibleSources\left(\anEvidence\right)}
    \weightConditional_{\idxDomainSource,\anEvidence} 
    \probabilityMeasure_{\domainSource_{\idxDomainSource}}\left(\anHypothesis \mid \anEvidence\right)
    \label{eq:arithmetic-weighted-mean-of-posteriors-v1}
\end{equation}
with
\begin{equation}
    \allPlausibleSources\left(\anEvidence\right)
    =
    \left\{
        \domainSource\in\allDomainsSource:
        \probabilityMeasure_{\domainSource}\left(\anEvidence\right)\ne 0
    \right\}
    \ne
    \emptyset
\end{equation}
and
\begin{equation}
     \weightConditional_{\idxDomainSource,\anEvidence}
     =
     \weightUnconditional_{\idxDomainSource} \frac{\probabilityMeasure_{\domainSource_{\idxDomainSource}}\left(\anEvidence\right)}{\probabilityMeasure_{\domainTarget}\left(\anEvidence\right)}
     =
     \frac{\weightUnconditional_{\idxDomainSource} \probabilityMeasure_{\domainSource_{\idxDomainSource}}\left(\anEvidence\right)}{\sum_{\idxDomainSource'=1}^{\numDomainSource} \weightUnconditional_{\idxDomainSource'} \probabilityMeasure_{\domainSource_{\idxDomainSource'}}\left(\anEvidence\right)}
     \dot
     \label{eq:conditional-weights-1}
\end{equation}
In the following, we pose  $\vectorWeightConditional_{\anEvidence}=(\weightConditional_{1,\anEvidence},\weightConditional_{2,\anEvidence}\ldots \weightConditional_{\numDomainSource,\anEvidence})$. From Equation~\ref{eq:conditional-weights-1}, we see that $\vectorWeightConditional_{\anEvidence}\in \simplexDomainSource$.

One recognizes in Equation~\ref{eq:arithmetic-weighted-mean-of-posteriors-v1} the \emph{distribution weighted combining rule} proposed by Mansour~\etal in~\cite{Mansour2008Domain}. The theoretical analysis performed in that paper shows that this combination rule behaves well when the posteriors in the source domains are affected by a bounded uncertainty.

\subsubsection{Motivation for determining $\vectorWeightConditional_{\anEvidence}$}
\label{subsubsec:without-conditional-weights}

\begin{figure}
\begin{centering}
\newcommand{\textA}{$\anHypothesis_1$}
\newcommand{\textB}{$\anHypothesis_2$}
\newcommand{\textC}{$\anHypothesis_3$}
\newcommand{\textD}{source domains}
\newcommand{\textE}{target domain}
\resizebox{1\linewidth}{!}{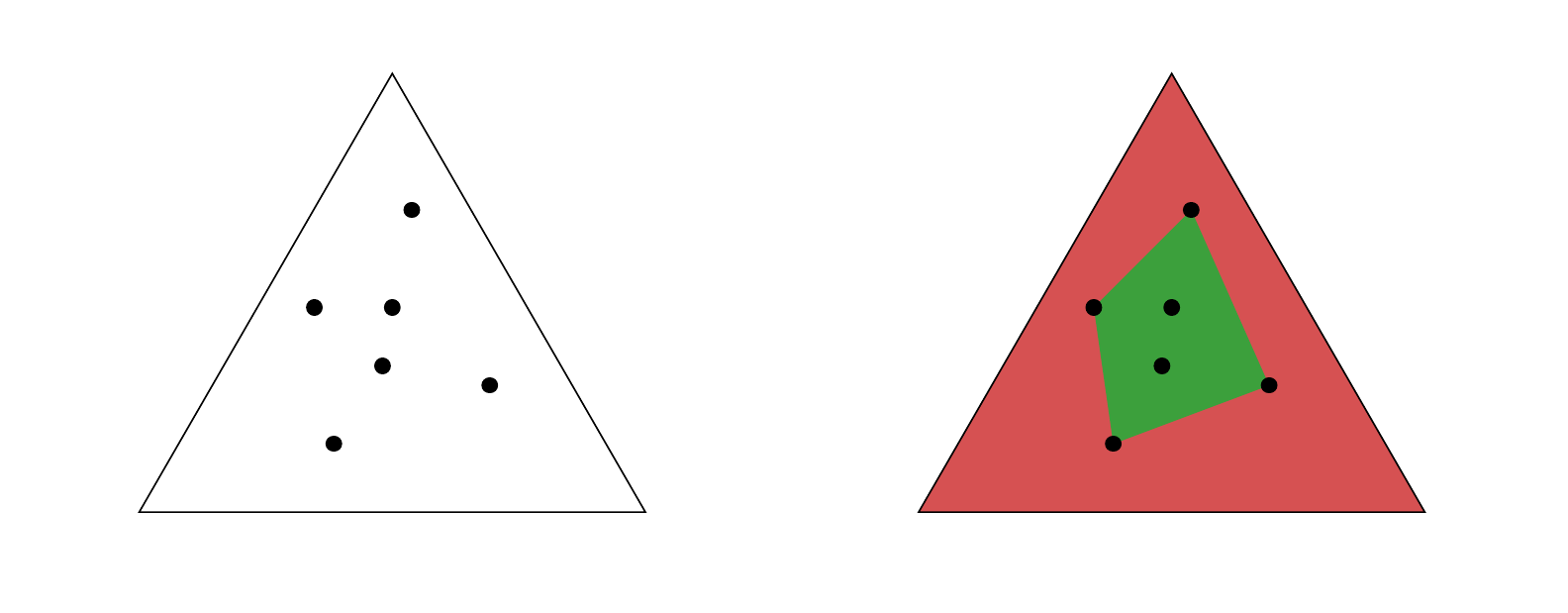}
\end{centering}
\caption{Given the vectors of posteriors in the source domains (black dots), there might be an important uncertainty on the vector of posteriors in the target domain (green area), as explained in Section~\ref{subsubsec:without-conditional-weights}. A solution is given in Section~\ref{subsubsec:with-conditional-weights}. In this figure, there are 6 source domains and $\allHypotheses=\left\{\anHypothesis_1,\anHypothesis_2,\anHypothesis_3\right\}$.}\label{fig:posteriors_in_triangle}
\end{figure}

In general, $\probabilityMeasure_{\domainTarget}\left(\anHypothesis \mid \anEvidence\right)$ cannot be determined based only on $\weightUnconditional_{\idxDomainSource}$ and $\probabilityMeasure_{\domainSource_{\idxDomainSource}}\left(\anHypothesis \mid \anEvidence\right)$, $\forall i\in\left\{1,2,…\numDomainSource\right\}$. This is because, unless some $\weightUnconditional_{\idxDomainSource}$ are null, the vector $\vectorWeightConditional_{\anEvidence}$ can sweep the complete probabilistic simplex $\simplexDomainSource$. 
Therefore, the posteriors for the target domain can be anywhere within the convex hull of the posteriors for the source domains $\domainSource_{\idxDomainSource}$ such that $\domainSource_{\idxDomainSource}\in\allPlausibleSources\left(\anEvidence\right)$ and $\weightUnconditional_{\idxDomainSource}>0$. Figure~\ref{fig:posteriors_in_triangle} shows this uncertainty for $\left|\allHypotheses\right|=3$.

Since MAP is independent of the priors, one can show, by a convexity argument, that, if the decisions are the same in all source domains, then it is also the same in the target domain. 
Computing $\weightConditional_{\anEvidence}$ is necessary only when the decisions taken in the source domains are contradictory.

On the contrary, since MLE depends on the priors, the decision made in the target domain can be other than the decisions made in the source domains. In particular, if the decisions are the same in all source domains, then the decision in the target domain could be different.

\subsubsection{Determination of $\vectorWeightConditional_{\anEvidence}$}
\label{subsubsec:with-conditional-weights}

In order to establish a way of computing $\vectorWeightConditional_{\anEvidence}$, we start by giving a probabilistic meaning to it. 
To this aim, we introduce the overall measurable space $\left(\sampleSpace',\eventSpace'\right)$ with $\sampleSpace'=\sampleSpace\times\allDomainsSource$ and $\eventSpace'=\eventSpace\times 2^{\allDomainsSource}$. We define the event $\eventDomainSource=\sampleSpace\times\{\domainSource_{\idxDomainSource}\}$ for the $\idxDomainSource$-th source domain and, for any event $\anEvent\in\eventSpace$, we define the corresponding event $\anEvent'=\anEvent\times\allDomainsSource\in\eventSpace'$.

All probabilities considered until now can be written in terms of a unique probability measure $\overallProbabilityMeasure$ on $\left(\sampleSpace',\eventSpace'\right)$. It is such that $
\overallProbabilityMeasure\left(\anEvent\times\{\domainSource_{\idxDomainSource}\}\right)
    =
    \weightUnconditional_{\idxDomainSource}
    \probabilityMeasure_{\domainSource_{\idxDomainSource}}\left(\anEvent\right)
$ for all $\anEvent\in\eventSpace$ and for all $\domainSource_{\idxDomainSource}\in\allDomainsSource$. For example, $\probabilityMeasure_{\domainSource_{\idxDomainSource}}\left(\anEvent\right)=\overallProbabilityMeasure\left(\anEvent'\mid\eventDomainSource\right)$ and $\probabilityMeasure_{\domainTarget}\left(\anEvent\right)=\overallProbabilityMeasure\left(\anEvent'\right)$, $\forall \anEvent\in\eventSpace$. Moreover, $\weightUnconditional_{\idxDomainSource}$ and $\vectorWeightConditional_{\anEvidence}$ acquire a probabilistic meaning as $\weightUnconditional_{\idxDomainSource}=\overallProbabilityMeasure\left(\eventDomainSource\right)$ and $\weightConditional_{\idxDomainSource,\anEvidence}=\overallProbabilityMeasure\left(\eventDomainSource\mid\anEvidence'\right)$.

Unfortunately, $\overallProbabilityMeasure$ cannot be used during the off-line stage because it depends on $\vectorWeightUnconditional$. Instead, we choose to work with the following probability measure $\overallProbabilityMeasure^{*}$ on $\left(\sampleSpace',\eventSpace'\right)$:
\begin{equation}
    \overallProbabilityMeasure^{*}\left(\anEvent\times\{\domainSource_{\idxDomainSource}\}\right)
    =
    \arbitrarilyChosenWeightUnconditional_{\idxDomainSource}
    \probabilityMeasure_{\domainSource_{\idxDomainSource}}\left(\anEvent\right)
    \comma
    \label{eq:probability-measure-to-train-source-disciminator}
\end{equation}
$\forall \anEvent\in\eventSpace$ and $\forall \domainSource_{\idxDomainSource}\in\allDomainsSource$, with any arbitrarily chosen vector of strictly positive weights $\vectorArbitrarilyChosenWeightUnconditional=(\arbitrarilyChosenWeightUnconditional_1,\arbitrarilyChosenWeightUnconditional_2\ldots \arbitrarilyChosenWeightUnconditional_{\numDomainSource})\in\simplexDomainSourceInterior$.\footnote{We use the notation $\simplexDomainSourceInterior$ to denote the interior of $\simplexDomainSource$, that is $\vectorArbitrarilyChosenWeightUnconditional\in\simplexDomainSourceInterior$ if and only if $\sum_{\idxDomainSource=1}^{\numDomainSource}\arbitrarilyChosenWeightUnconditional_\idxDomainSource=1$ and $\arbitrarilyChosenWeightUnconditional_{\idxDomainSource}>0 \, \forall \idxDomainSource$.}

In the same way that we assumed the existence of a model determining $\probabilityMeasure_{\domainSource}\left(\anHypothesis\mid\anEvidence\right)$ for any $\anEvidence\in\allEvidences$ and any $\anHypothesis\in\allHypotheses$, we can also assume that it is possible to obtain a model for $\overallProbabilityMeasure^{*}\left(\eventDomainSource\mid\anEvidence'\right)$, for any given $\anEvidence\in\allEvidences$ and any given source domain.  
As $\overallProbabilityMeasure^{*}\left(\eventDomainSource\right)=\arbitrarilyChosenWeightUnconditional_{\idxDomainSource}$ and $\overallProbabilityMeasure^{*}\left(\anEvidence'\mid\eventDomainSource\right)=\probabilityMeasure_{\domainSource_{\idxDomainSource}}\left(\anEvidence\right)\,\forall\anEvidence\in\allEvidences$, Equation~\ref{eq:conditional-weights-1} leads to
\begin{equation}
    \weightConditional_{\idxDomainSource,\anEvidence}
    =
    \frac{
        \frac{
            \vectorWeightUnconditional_{\idxDomainSource}
        }{
            \vectorArbitrarilyChosenWeightUnconditional_{\idxDomainSource}
        }
        \,
        \overallProbabilityMeasure^{*}\left(\eventDomainSource\mid\anEvidence'\right)
    }{
        \sum_{\idxDomainSource'=1}^{\numDomainSource}
        \frac{
            \vectorWeightUnconditional_{\idxDomainSource'}
        }{
            \vectorArbitrarilyChosenWeightUnconditional_{\idxDomainSource'}
        }
        \,
        \overallProbabilityMeasure^{*}\left(\eventDomainSource[\idxDomainSource']\mid\anEvidence'\right)
    }
    \dot
    \label{eq:conditional-weights-2}
\end{equation}
This equation is similar to Equation~\ref{eq:target-shift} and can be interpreted as a prior shift for priors on the $\numDomainSource$ source domains instead of priors on the $\left|\allHypotheses\right|$ hypotheses.

\subsection{Our domain adaptation algorithm}
\label{subsec:our-algorithm}

\begin{figure}
\begin{centering}
\newcommand{\textA}{domain $\domainSource_1$}
\newcommand{\textB}{source model}
\newcommand{\textC}{domain $\domainSource_2$}
\newcommand{\textD}{source model}
\newcommand{\textE}{domain $\domainSource_\numDomainSource$}
\newcommand{\textF}{evidence $\anEvidence$}
\newcommand{\textG}{target shift}
\newcommand{\textH}{(see Equation~\ref{eq:target-shift})}
\newcommand{\textI}{target shift}
\newcommand{\textJ}{(see Equation~\ref{eq:target-shift})}
\newcommand{\textK}{target shift}
\newcommand{\textL}{target shift}
\newcommand{\textM}{(see Equation~\ref{eq:conditional-weights-2})}
\newcommand{\textN}{(see Equation~\ref{eq:target-shift})}
\newcommand{\textO}{source model}
\newcommand{\textP}{(\wrt Equation~\ref{eq:probability-measure-to-train-source-disciminator})}
\newcommand{\textQ}{dom. discri. model}
\newcommand{\textR}{$\probabilityMeasure_{\domainSource_1}\left(\anHypothesis | \anEvidence\right)$}
\newcommand{\textS}{$\probabilityMeasure_{\domainSource_2}\left(\anHypothesis | \anEvidence\right)$}
\newcommand{\textT}{$\probabilityMeasure_{\domainSource_\numDomainSource}\left(\anHypothesis | \anEvidence\right)$}
\newcommand{\textU}{posteriors $\probabilityMeasure_{\domainTarget}\left(\anHypothesis | \anEvidence\right)$}
\newcommand{\textV}{$\weightConditional_{\anEvidence}$}
\newcommand{\textW}{weighted arithmetic mean}
\newcommand{\textX}{(see Equation~\ref{eq:arithmetic-weighted-mean-of-posteriors-v1})}
\newcommand{\textZ}{$\overallProbabilityMeasure^{*}\left(\eventDomainSource | \anEvidence'\right)$}
\newcommand{\textAA}{$\probabilityMeasure^{*}_{\domainSource_1}\left(\anHypothesis | \anEvidence\right)$}
\newcommand{\textAB}{$\probabilityMeasure^{*}_{\domainSource_2}\left(\anHypothesis | \anEvidence\right)$}
\newcommand{\textAC}{$\probabilityMeasure^{*}_{\domainSource_\numDomainSource}\left(\anHypothesis | \anEvidence\right)$}
\resizebox{1\linewidth}{!}{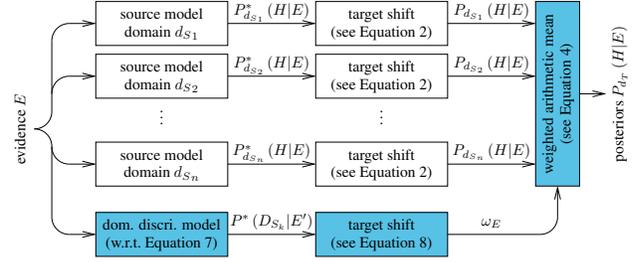}
\end{centering}
\caption{Our unsupervised domain adaptation algorithm. The boxes with a white background are used for determining the posteriors in the source domains. Those with a blue background depict the elements that we have added to obtain, on the fly, the posteriors in the target domain.
}\label{fig:algorithm}
\end{figure}

Figure~\ref{fig:algorithm} shows the algorithm that we propose to compute $\probabilityMeasure_\domainTarget\left(\anHypothesis\mid\anEvidence\right)$ on the fly for any hypothesis $\anHypothesis\in\allHypotheses$, any evidence $\anEvidence\in\allEvidences$, and any target domain $\domainTarget\in\allDomainsTarget$. 

\mysection{The off-line step.}
Two types of models are obtained off-line, before knowing the target domain: $\left|\allDomainsSource\right|$ \emph{source models} and 1 \emph{domain discriminator model}. They are used to determine $\probabilityMeasure^{*}_{\domainSource_{\idxDomainSource}}\left(\anHypothesis\mid\anEvidence\right)$ and $\overallProbabilityMeasure^{*}\left(\eventDomainSource\mid\anEvidence'\right)$, respectively.
Training the source models requires single-source labeled data. Training the domain discriminator model requires multiple-source unlabeled data. Equations~\ref{eq:target-shift-hypothesis} and~\ref{eq:probability-measure-to-train-source-disciminator} specify the distributions that have to be assumed by the learner.

\mysection{The on-the-fly step.}
Equation~\ref{eq:target-shift} is applied to obtain $\probabilityMeasure_{\domainSource_{\idxDomainSource}}\left(\anHypothesis\mid\anEvidence\right)$ from $\probabilityMeasure^{*}_{\domainSource_{\idxDomainSource}}\left(\anHypothesis\mid\anEvidence\right)$.
Moreover, knowing the target domain (\ie the vector $\vectorWeightUnconditional$), Equation~\ref{eq:conditional-weights-2} is applied to obtain $\weightConditional_{\idxDomainSource,\anEvidence}$ from $\overallProbabilityMeasure^{*}\left(\eventDomainSource\mid\anEvidence'\right)$.
Finally, Equation~\ref{eq:arithmetic-weighted-mean-of-posteriors-v1} leads to the posterior $\probabilityMeasure_\domainTarget\left(\anHypothesis\mid\anEvidence\right)$ in the target domain.
The overall algorithm gives the flexibility to adapt the posteriors on the fly to an evolving target domain.

\mysection{Discussion.}
If all the source models and the domain discriminator model are \emph{exact up to a target shift}, then our algorithm behaves as an \emph{exact model} for the target domain. In practice, however, models are rarely exact (\ie they have noisy outputs). Taking the $L_1$ distance to measure the errors, it can be shown that: if the error on $\weightConditional_{\anEvidence}$ (over $\allDomainsSource$) is bounded by $\noiseDomainDiscriminatorModel$ and if the error on $\probabilityMeasure_{\domainSource}\left(\anHypothesis\mid\anEvidence\right)$ (over $\allHypotheses$, a partition of $\sampleSpace$) is bounded by $\noiseSourceModels$ for all $\domainSource\in\allDomainsSource$, then the error on $\probabilityMeasure_{\domainTarget}\left(\anHypothesis\mid\anEvidence\right)$ is bounded by $\noiseOutput\le\noiseSourceModels+\noiseDomainDiscriminatorModel$.

In other words, the last step of the algorithm behaves well when facing noisy inputs. However, theoretically, the upstream target shift operations can amplify the noise affecting the outputs of the various models, especially when priors are low. Therefore, in the next section, we will demonstrate the practical suitability of this algorithm, in the challenging case in which some priors are low. Also, we validate the predicted posteriors after the target shifts.

\section{Experiments\label{sec:experiments}}


\subsection{Experimental Setup}
\label{subsec:ExperimentalSetup}
We evaluate and compare our domain adaptation algorithm on the semantic segmentation task, which consists in predicting, for each pixel of an image, the semantic class of its enclosing object or region in the scene.

  In accordance with the previous section, the notations are as follows. We denote the set of images by $\allImages$, the set of pixels by $\allPixels$, and the set of semantic classes by $\allSemanticClasses$. All these sets being finite in our experimental setup, we take $\sampleSpace = \allImages \times \allPixels \times \allSemanticClasses$ and $\eventSpace=2^\sampleSpace$. The evidence $\anEvidence_{\anImage,\aPixel}$ for the pixel $\aPixel$ in image $\anImage$ is $\left\{(\anImage,\aPixel,\aSemanticClass)\mid\aSemanticClass\in\allSemanticClasses\right\}$ and the hypothesis $\anHypothesis_{\aSemanticClass}$ for the groundtruth semantic class $\aSemanticClass$ is $\left\{(\anImage,\aPixel,\aSemanticClass)\mid\anImage\in\allImages,\,\aPixel\in\allPixels\right\}$.  Note that $\allHypotheses$ is a partition of $\sampleSpace$. With our notations, the semantic segmentation task aims at choosing an element of $\allSemanticClasses$, the predicted semantic class, based on the posteriors $\probabilityMeasure_{\domain}\left(\anHypothesis_{\aSemanticClass}\mid\anEvidence_{\anImage,\aPixel}\right)$.

\subsubsection{Off-line choices}

Our algorithm requires two types of off-line models: (1)~source models and (2)~a domain discriminator model. In the following, we describe each choice for these models.

\mysection{Choice of models.}
Various machine learning techniques can be used as posterior estimators. In general, deep learning is suitable for predicting posteriors when the loss minimized during the training stage is carefully chosen. In particular, the cross-entropy reaches a local minimum when the output of the network corresponds to the posteriors $\probabilityMeasure_{\domain}\left(\anHypothesis\mid\anEvidence\right)\,\forall\anHypothesis\in\allHypotheses$~\cite{Miller1993OnLoss,Ramos2018Deconstructing}. 
For the architecture of the source models and the domain discriminator model, we chose the TinyNet~\cite{Cioppa2019ARTHuS,Cioppa2018ABottomUp} segmentation network, which is a lightweight architecture that only needs a few training samples and is fast to train.

\mysection{Training the source models.}
For all our experiments, we train each source model on its own source dataset separately. 
The models are trained with batches of $12$ images randomly sampled from the training set of the source domain.
We use a learning rate of $10^{-4}$ with a reduce-on-plateau-scheduling strategy, a patience of $10$ and a reduction factor of $0.1$.
We use the Adam optimizer with default parameters and no weight decay~\cite{Kingma2015Adam}.
To avoid typical problems when dealing with unbalanced distributions of classes, we use the weighted cross-entropy loss, for which the weighting factor is estimated using the priors of the source domain. These choices are motivated by the wish to obtain source models \emph{exact up to a target shift}. We aim at satisfying Equation~\ref{eq:target-shift-hypothesis} with $\probabilityMeasure^{*}_{\domainSource}\left(\anHypothesis_{\aSemanticClass}\right)=\left|\allSemanticClasses\right|^{-1}\,\forall \anHypothesis_{\aSemanticClass}\in\allHypotheses$. Indeed, the chosen loss achieves a local minimum for a pixel $\aPixel$ in an image $\anImage$ when the source model $\functionSourceModel:\allEvidences\rightarrow\simplexSemanticClasses$ gives $\functionSourceModel(\anImage,\aPixel)_\aSemanticClass=\probabilityMeasure^{*}_{\domainSource}\left(\anHypothesis_{\aSemanticClass}\mid\anEvidence_{\anImage,\aPixel}\right)$ for all semantic classes (indexed by $\aSemanticClass$ here).

\mysection{Training the domain discriminator model.}
To estimate $\weightConditional_{\idxDomainSource,\anEvidence}$, we train a soft discriminator to recognize the corresponding domain of each pixel. We use the same TinyNet architecture, training procedure, and hyperparameters than for training the source models. Nevertheless, we replace the number of semantic classes with the number of source domains in the output layer and the batch size by $4$.

Regarding training samples, we propose a way to generate multi-domain images, inspired by the mosaic transformation presented in~\cite{Bochkovskiy2020YOLOv4}. We combine four patches cropped from four randomly drawn images to create new training images. 
These choices are motivated by the wish to satisfy Equation~\ref{eq:probability-measure-to-train-source-disciminator} with $\arbitrarilyChosenWeightUnconditional_\idxDomainSource=\left|\allDomainsSource\right|^{-1}$. Indeed, the cross-entropy loss achieves a local minimum for a pixel $\aPixel$ in an image $\anImage$ when the domain discriminator model $\functionDomainDiscriminatorModel:\allEvidences\rightarrow\simplexDomainSource$ gives $\functionDomainDiscriminatorModel(\anImage,\aPixel)_\idxDomainSource=\overallProbabilityMeasure^{*}\left(\eventDomainSource\mid\anEvidence'_{\anImage,\aPixel}\right)$ for all source domains (indexed by $\idxDomainSource$ here).

\mysection{Validation of predicted posteriors.}
The source models and the domain discriminator model are expected to be exact up to a target shift. Throughout our experiments, we systematically perform two tests to establish their trustablility. First, we compute the posteriors estimated by these models and verify (after the necessary target shifts)  that these predictions correspond, in expectation, to the respective priors. Second, we verify by visual inspection that these models are well calibrated using calibration plots following~\cite{NiculescuMizil2005Predicting,scikit-learn}.

\subsubsection {Evaluation and comparison with heuristics}

For the sake of evaluation, we simulate various target domains by mixing the test sets of the different source domains. This is achieved by weighting the images. In our experiments, all images have the same size and all test sets contain the same amount of images. Thus, we weight the images of the $\idxDomainSource$-th test set by $\weightUnconditional_\idxDomainSource$ to satisfy Equation~\ref{eq:domain-target-fct-sources}.

We use $4$ common performance scores to evaluate the quality of the semantic segmentations: the accuracy, balanced accuracy, mean IoU (macro-averaging), and balanced mean IoU. All reported results are for the decision-making strategy MAP. We compare the scores obtained with our algorithm to those obtained with the $3$ following heuristics.

\begin{enumerate}
\item \textbf{Source models.} The first heuristic consists in using each source model separately on the target domains. For a fair comparison, we apply a target shift on the posteriors corresponding to the target domain priors following Equation~\ref{eq:target-shift}.

\item \textbf{Random selection of source models.}
The second heuristic randomly selects the decision derived from the $\idxDomainSource$-th  source model with a probability given by $\weightUnconditional_\idxDomainSource$. 

\item \textbf{Linear combination of posteriors.}
The third heuristic combines linearly the posteriors provided by the source models as follows: 
$    
    \sum_{\idxDomainSource=1}^{\numDomainSource}
    \weightUnconditional_{\idxDomainSource}\,
    \probabilityMeasure_{\domainSource_{\idxDomainSource}}\left(\anHypothesis_{\aSemanticClass}\mid\anEvidence_{\anImage,\aPixel}\right)
$. 

\end{enumerate}


\subsection{Experiment with 2 source domains}


In the first experiment, we consider the semantic segmentation of high-definition ($1280 \times 720$) color images acquired by cameras installed in moving vehicles, behind the windshield. In this experiment, there are $2$ source domains and $19$ strongly imbalanced semantic classes.

\mysection{Motivation.}
The semantic segmentation of such images is reputed to be a preliminary step towards autonomous vehicles.
There is a native need for considering multiple domains as we expect differences in the appearance of roads, traffic signs, and cars from one country to another. Moreover, the sensors and their positioning can differ from one car to another. And, last but not least, the weather and traffic conditions can vary continuously.
The authors of~\cite{Yu2020BDD100K} noted a dramatic domain shift between two datasets (BDD100K~\cite{Yu2020BDD100K} and Cityscapes~\cite{Cordts2016The}). According to their results, the semantic segmentation models perform much worse when tested on a different dataset.

\begin{figure}
\begin{centering}
\,\hfill{}\includegraphics[width=0.45\linewidth]{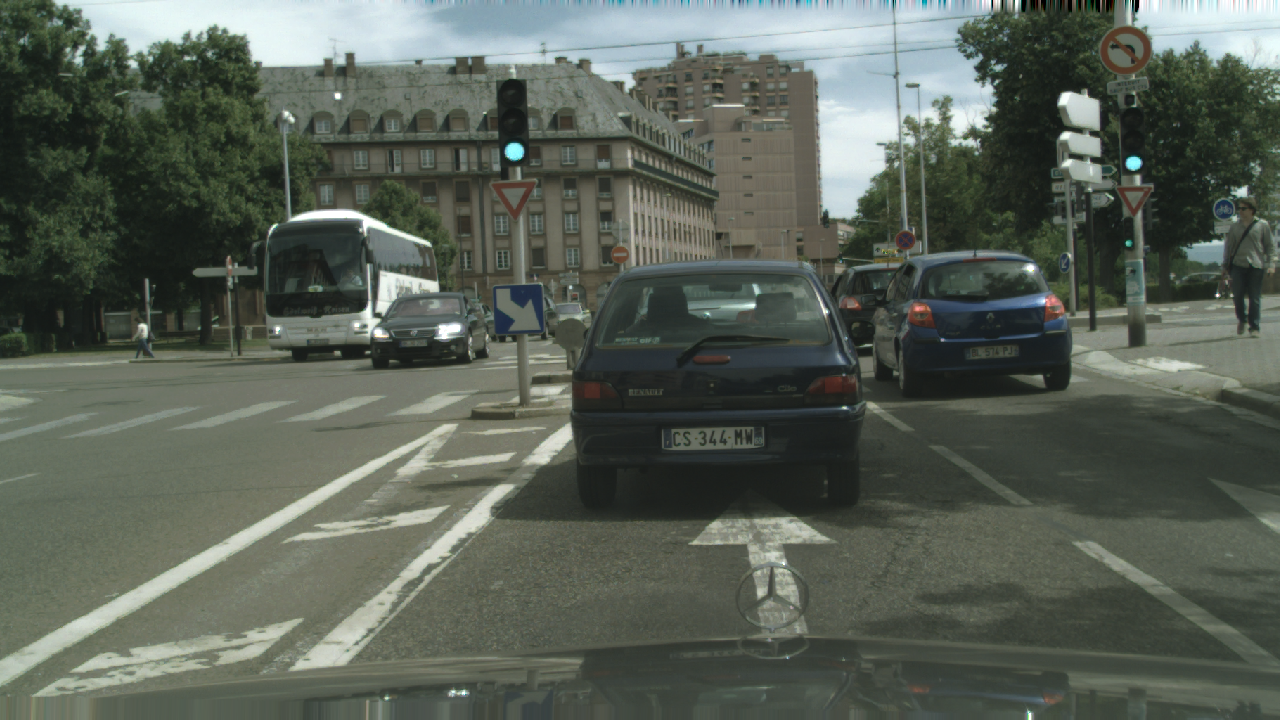}\hfill{}\includegraphics[width=0.45\linewidth]{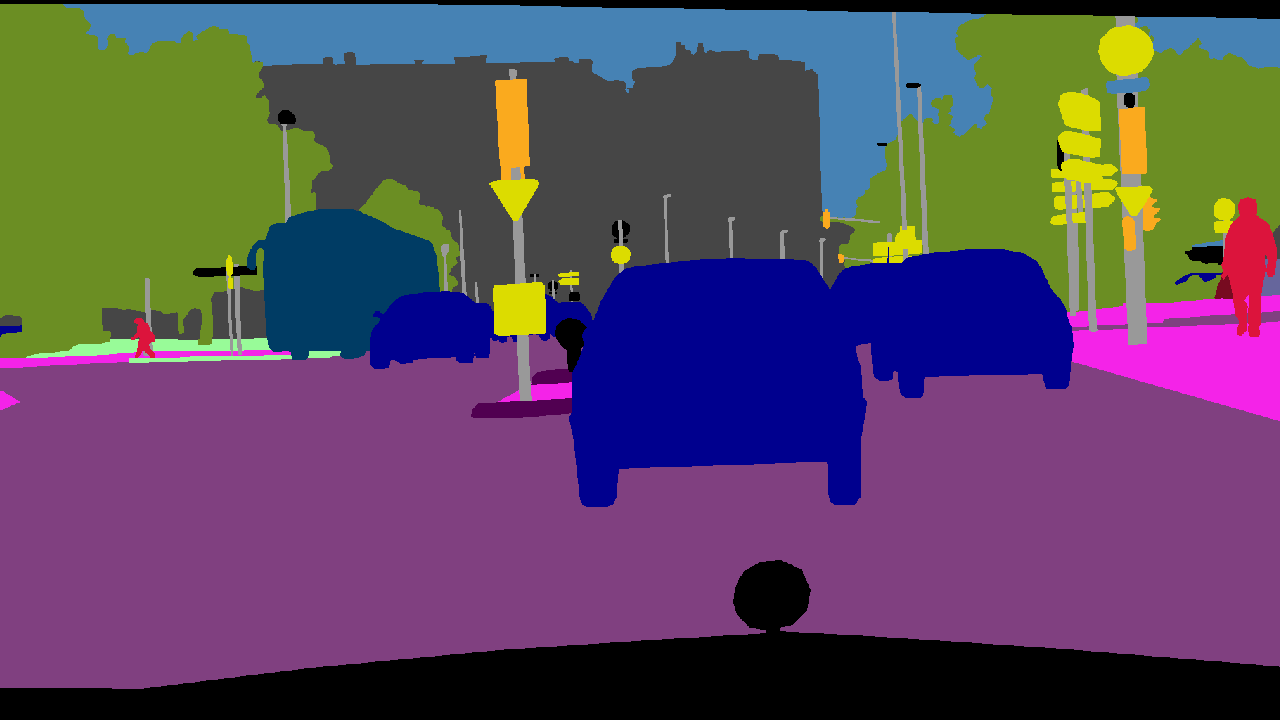}\hfill{}\,\\
\,\hfill{}\includegraphics[width=0.45\linewidth]{figures/dataset_BDD100K_city_input.jpg}\hfill{}\includegraphics[width=0.45\linewidth]{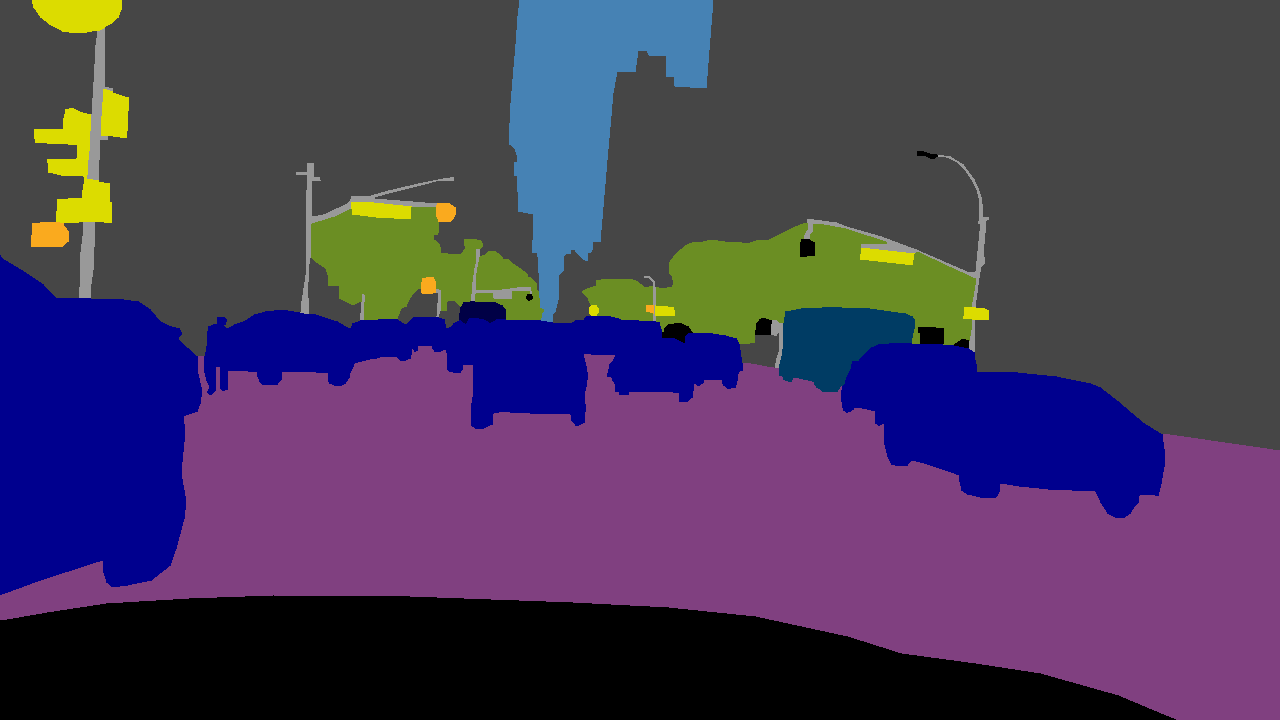}\hfill{}\,\\
\par\end{centering}
\caption{Examples of images (left) from the datasets Cityscapes (above) and BDD100K (below), with the corresponding groundtruth images (right).\label{fig:images-CityScapes-BDD100k}}
\end{figure}

\mysection{Data.} The source domains are represented by the datasets CityScapes (data acquired in European cities, mostly in Germany) and BDD100K
(data acquired in the USA). These datasets gather color images with their respective semantic segmentation groundtruths (see Figure~\ref{fig:images-CityScapes-BDD100k}). We resized and cropped all images to $1280 \times 720$.
We also randomly split each dataset into a training set (2726 images per dataset), a validation set (250 images per dataset), and a test set (500 images per dataset).

\begin{figure}
\begin{centering}
\includegraphics[width=1.0\linewidth]{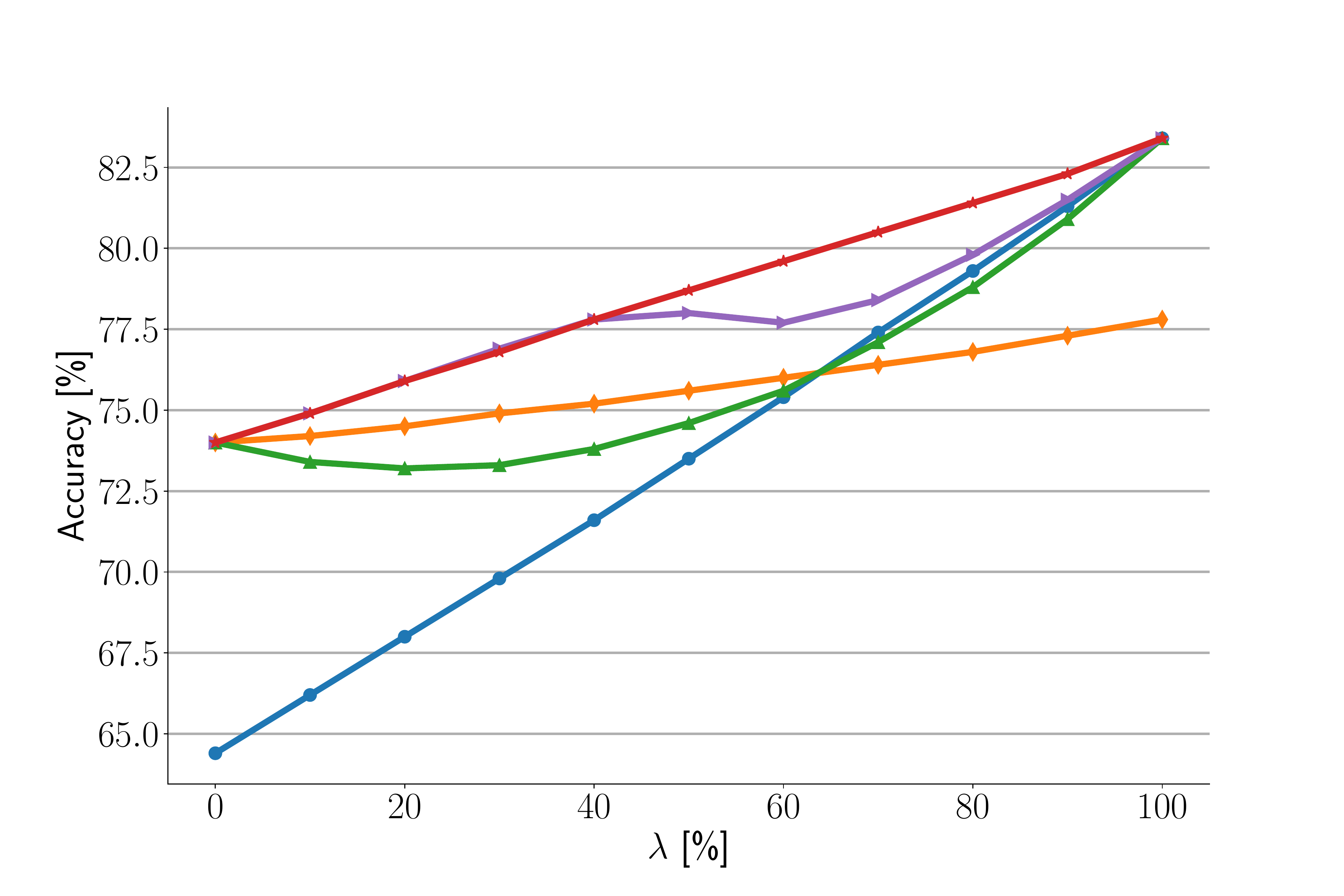}\\
\end{centering}
\caption{Results of our first experiment (all decisions made with the MAP strategy) showing the accuracy with the \myblue{source model trained on CityScapes (in blue)}, the \myorange{source model trained on BDD100K (in orange)}, the \mygreen{random selection of source models (in green)}, the \mypurple{linear combination of posteriors (in purple)}, and \myred{our algorithm (in red)}. $\weightUnconditional=0$ (resp. $=100)$ corresponds to BDD100K (resp. CityScapes) as target domain.}\label{fig:driving_accuracy}
\end{figure}

\begin{figure}
\begin{centering}
\,\hfill{}\includegraphics[width=0.45\linewidth]{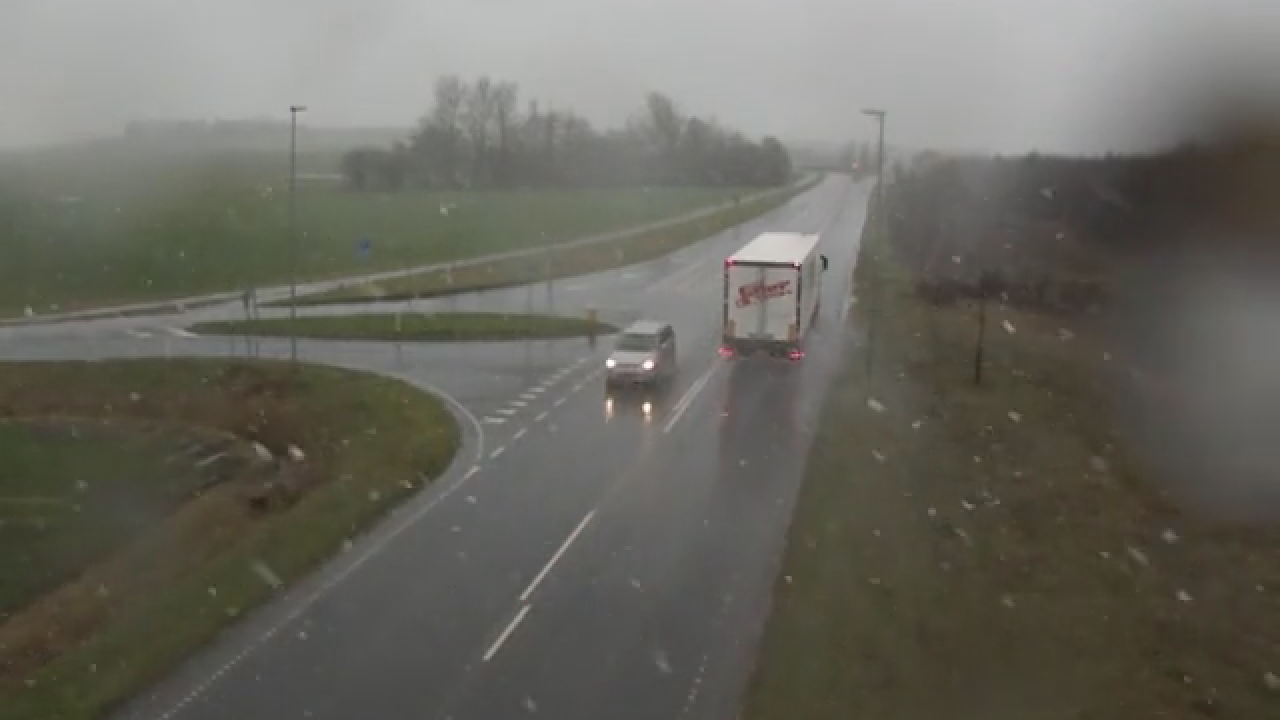}\hfill{}\includegraphics[width=0.45\linewidth]{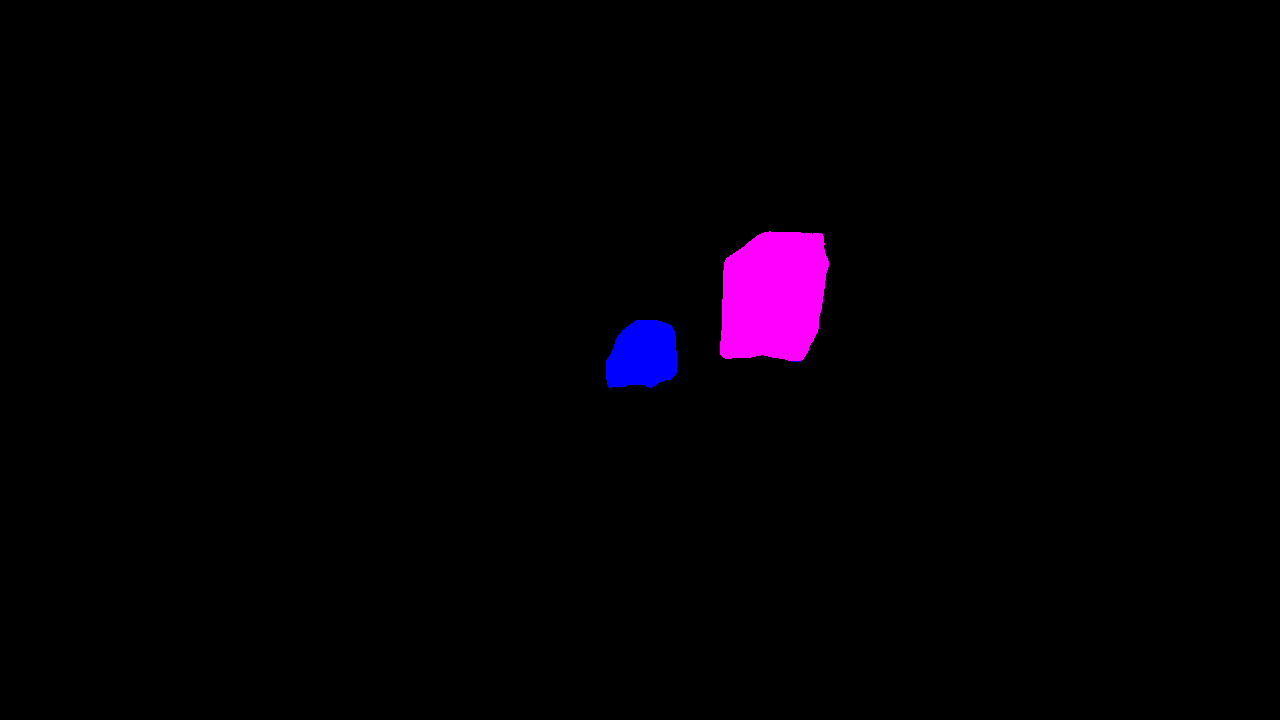}\hfill\,\\
\,\hfill{}\includegraphics[width=0.45\linewidth]{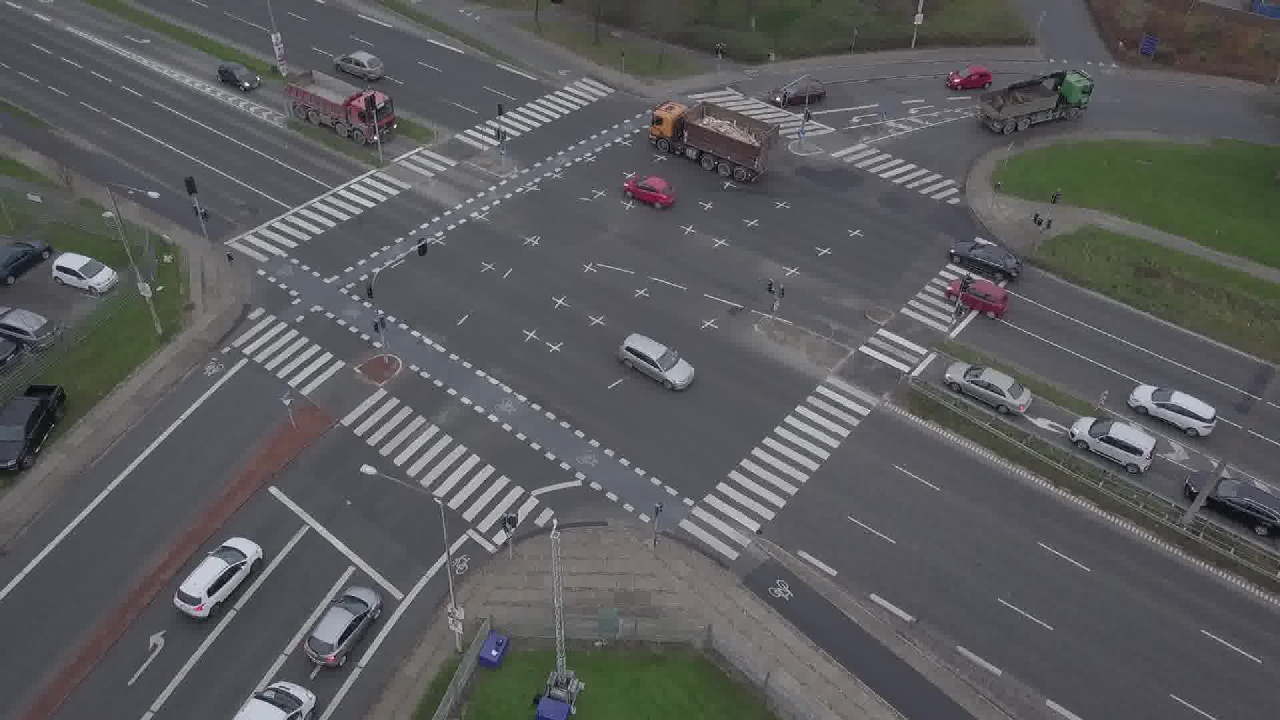}\hfill{}\includegraphics[width=0.45\linewidth]{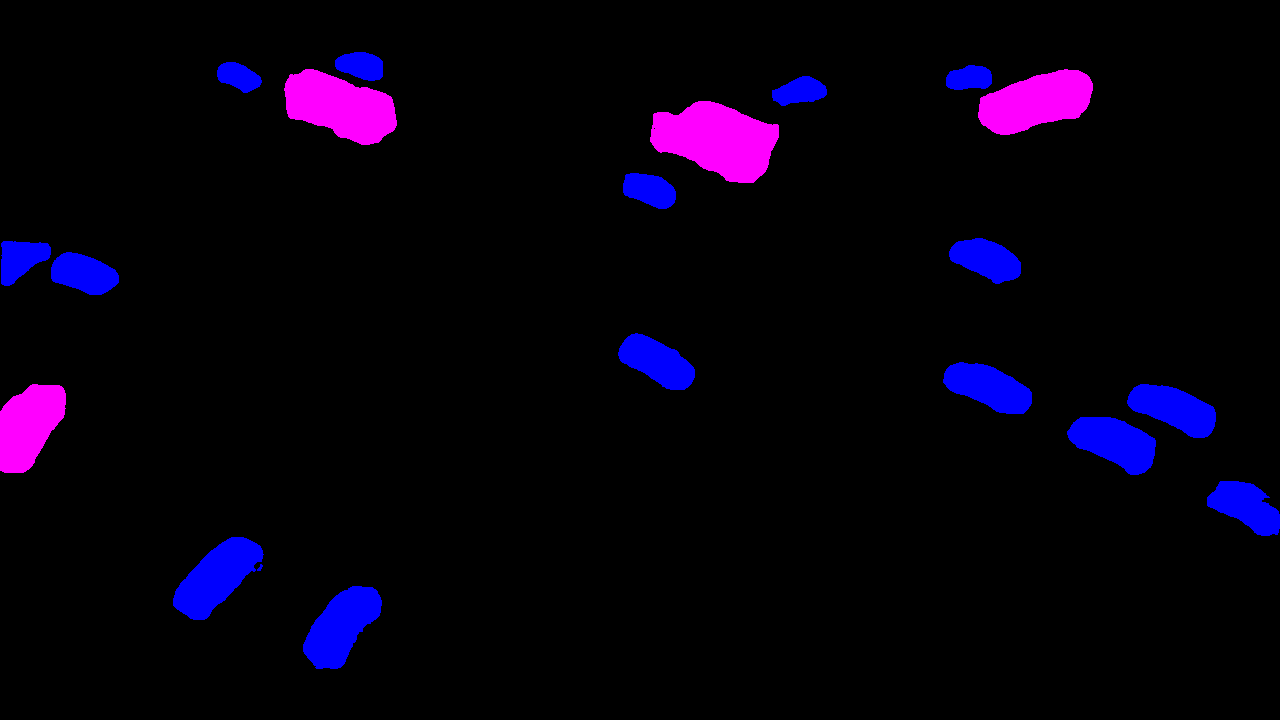}\hfill\,\\
\,\hfill{}\includegraphics[width=0.45\linewidth]{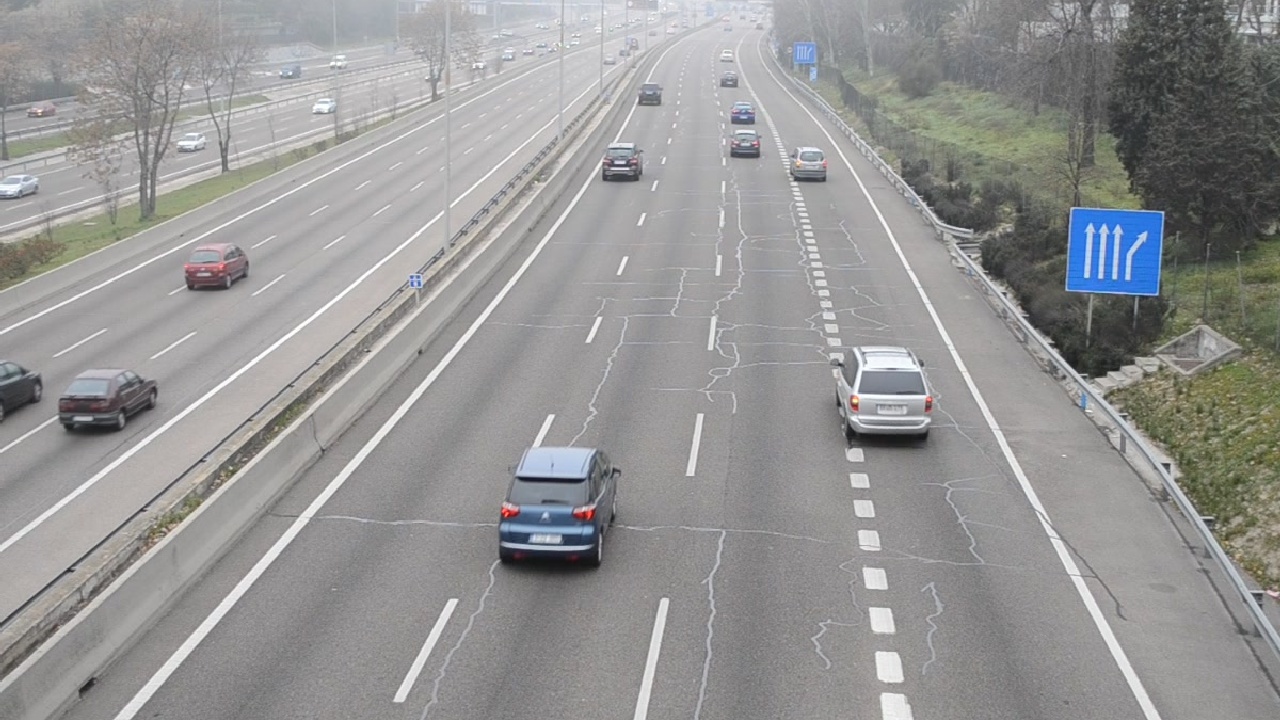}\hfill{}\includegraphics[width=0.45\linewidth]{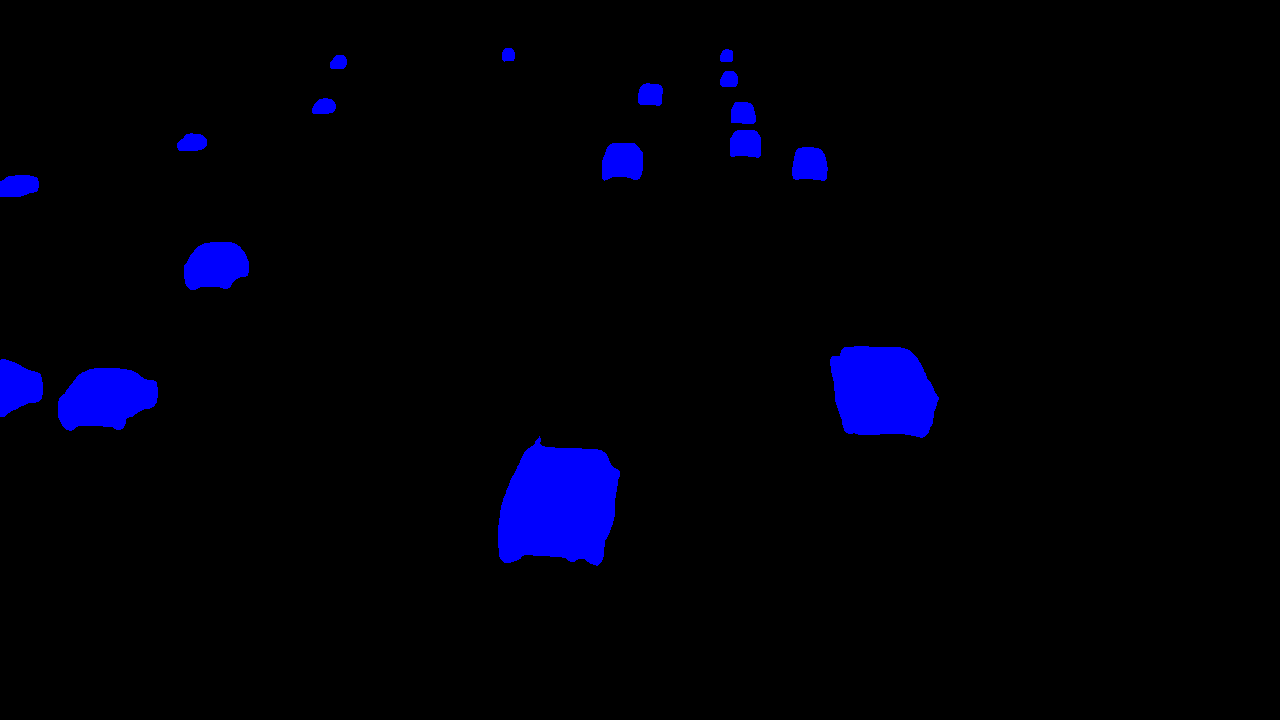}\hfill{}\,\\
\,\hfill{}\includegraphics[width=0.45\linewidth]{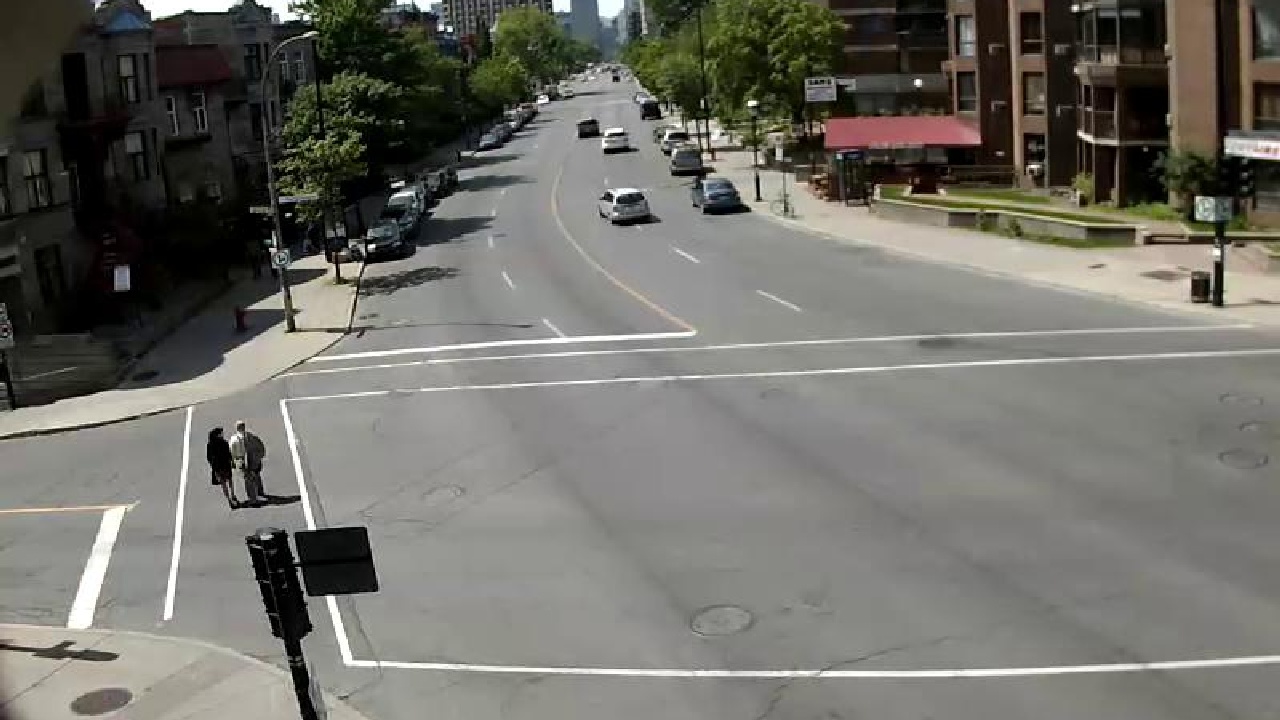}\hfill{}\includegraphics[width=0.45\linewidth]{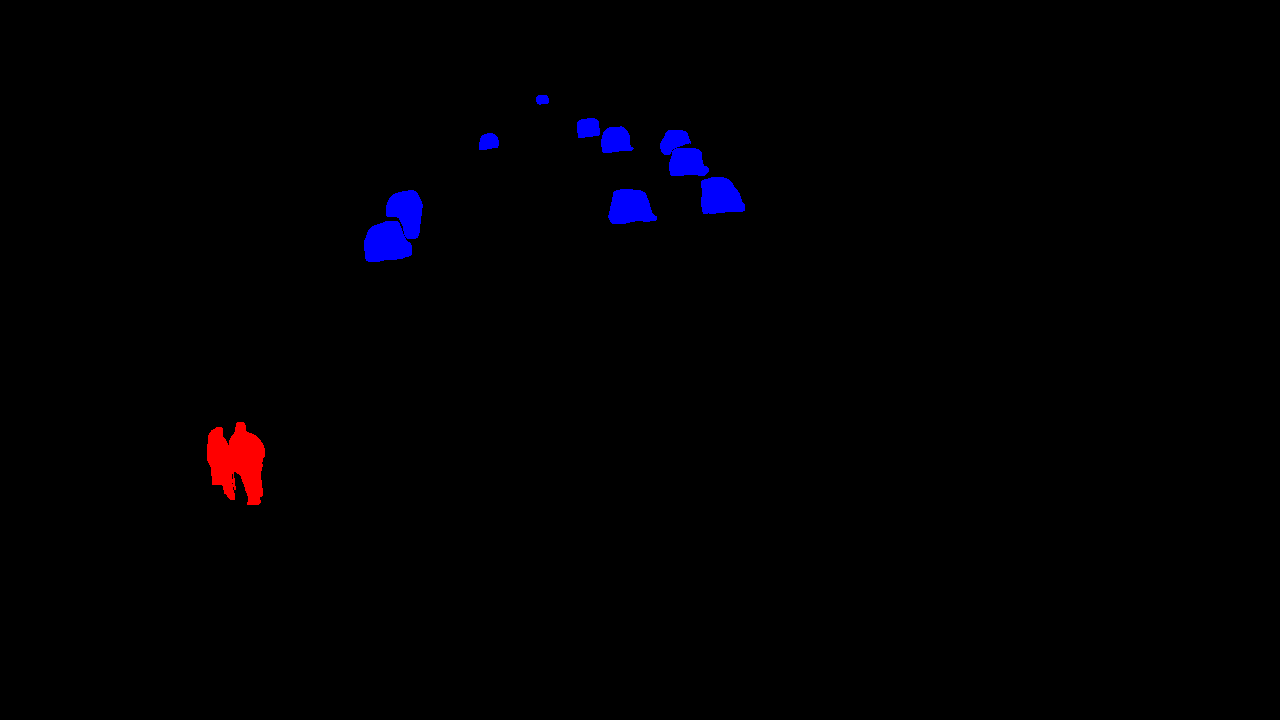}\hfill{}\,\\
\par\end{centering}
\caption{Examples of images (left) from the surveillance datasets (from top to bottom) RainSnow, MTID/Drone, GRAM-RTM/M-30-HD and UT/Sherbrooke, with the corresponding pseudo-groundtruth images (right).\label{fig:images-Surveillance}}
\end{figure}

\begin{figure*}
\begin{centering}
\,\hfill{}\includegraphics[width=0.24\linewidth]{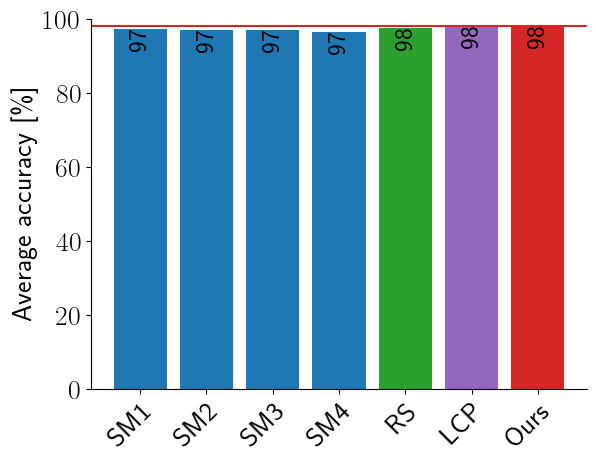}\hfill{}
\includegraphics[width=0.24\linewidth]{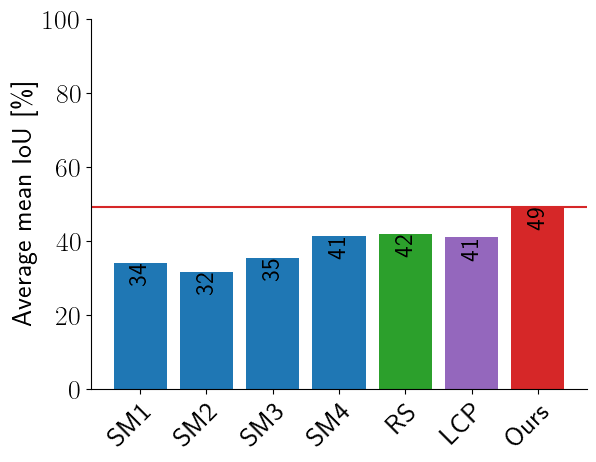}\hfill{}
\includegraphics[width=0.24\linewidth]{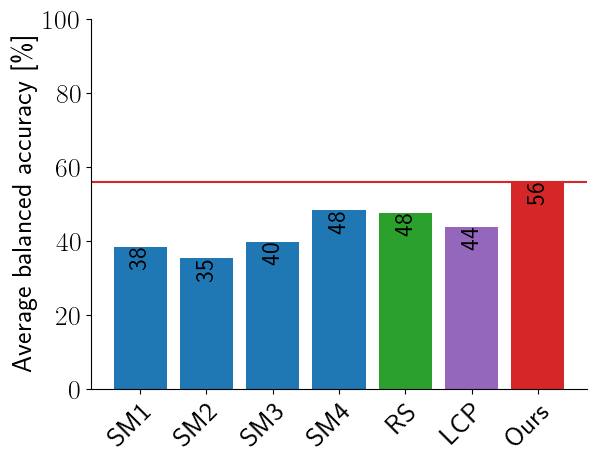}\hfill{}
\includegraphics[width=0.24\linewidth]{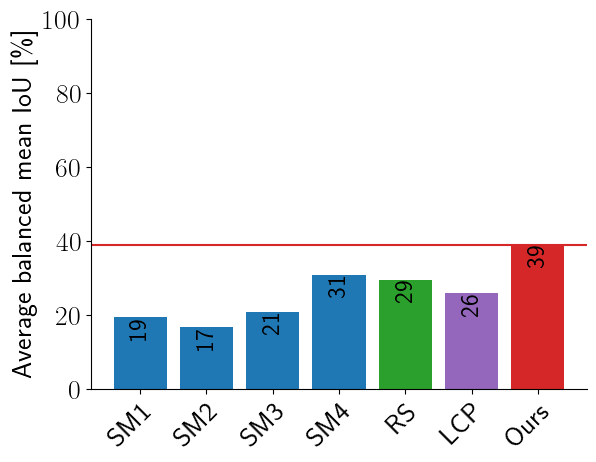}\hfill{}\,\\
\par\end{centering}
\caption{Results of our second experiment (all decisions made with the MAP strategy) showing the mean behaviour  over 15 diversified target domains. We compare the different \myblue{source models (SM1-4, in blue)} with the \mygreen{random selection of source models (RS, in green)}, \mypurple{linear
combination of posteriors (LCP, in purple)}, and \myred{our algorithm (in red)}. Four performance scores are averaged over the 15 target domains: the accuracy, mean IoU, balanced accuracy, and balanced mean IoU (from left to right). \label{fig:results-4sources}}
\end{figure*}

\mysection{Results.}
We compare the accuracy of our algorithm with the $3$ heuristics in Figure~\ref{fig:driving_accuracy}.

The first heuristic (source models) leads to the orange (model trained on BDD100K) and blue (model trained on CityScapes) curves. When the target domain coincides with one of the source domains (both sides of the graph), the corresponding source model behaves much better than the other one. This confirms the observation made in~\cite{Yu2020BDD100K} and the need for domain adaptation, as simply choosing a source model and applying it directly to a different target domain leads to a drastic decrease in performance.

The second heuristic (random selection of source models, in green) is clearly suboptimal compared to the best source model for the current target domain. It is therefore not an effective way to combine the source models.

The third heuristic (linear combination of the posteriors, in purple) reaches better performances than the two previous heuristics. This is a surprise as this simple heuristic is known to be inappropriate, by theoretical arguments~\cite{Mansour2008Domain}. 

Our domain adaptation algorithm (red curve) outperforms all heuristics, especially when the weight of CityScapes is above $40\%$ in the target domain mixture.


\subsection{Experiment with 4 source domains}

Let us consider a second  semantic segmentation task, with images acquired by fixed video surveillance traffic cameras. In this experiment, there are $4$ source domains and $4$ strongly imbalanced semantic classes.

\mysection{Data.} The four source domains are represented by the following datasets. (1) AAU RainSnow Traffic Surveillance Dataset~\cite{Bahnsen2019Rain} (called \textit{RainSnow}) contains traffic surveillance videos in rainfall and snowfall from seven different intersections. 
(2) Multi-view Traffic Intersection Dataset (MTID)~\cite{Jensen2020Presenting} (called \textit{MTID/Drone}) contains footage of the same intersection from two different points of view. We only use the images recorded with a drone from one point of view. 
(3) GRAM Road-Traffic Monitoring (GRAM-RTM) dataset~\cite{Guerrero2013Vehicle}  (called \textit{GRAM-RTM/M-30-HD}) consists of three videos recorded under different conditions and with different platforms. We only use one video, named M-30-HD. (4) The dataset from the Urban Tracker~\cite{Jodoin2014Urban} project (called \textit{UT/Sherbrooke}) contains also several videos. We only select the Sherbrooke video, filmed at the Sherbrooke/Amherst intersection in Montreal.

For each dataset, we randomly select $1299$ images for the training set, $300$ images for the validation set and $300$ images for the test set.
All images have been resized (\eg by upsampling, downsampling, or cropping) to $1280 \times 720$. 

Since no segmentation groundtruth is available for all of these datasets and since we need all datasets to share the same semantic classes, we use the PointRend~\cite{Kirillov2020PointRend} algorithm trained on the COCO~\cite{Lin2014COCO} dataset to obtain pseudo-groundtruths. For this experiment, we consider $4$ semantic classes: background, person, two wheels (bicycle and motorcycle), and four wheels (car, bus, and truck). 
An example of an image and its pseudo-groundtruth for each dataset is illustrated in Figure~\ref{fig:images-Surveillance}.

\mysection{Performance analysis.} Since we have $4$ source domains in this experiment, it is impossible to depict the performances as we did in the first experiment.  Instead, Figure~\ref{fig:results-4sources} shows the mean behaviour over several target domains. Bar plots are provided for $4$ performance scores. They are obtained by averaging the  scores of our algorithm and of the heuristics over $15$ target domains obtained by mixing either $1$, $2$, $3$, or $4$ source domains with equal weights.
As can be seen, on average, our algorithm outperforms all heuristics, with a large margin for three of the four scores. This demonstrates the superiority of our algorithm.

\section{Conclusion\label{sec:conclusion}}

In this paper, we focus on the domain adaptation problem in which the probability measure of the target domain is a convex combination of the probability measures of the source domains.
We define a probabilistic framework and show that the posteriors in the target domain do not depend solely on posteriors, priors, and relative weights of the source domains.
From there, we provide a theoretical proof of an algorithm to compute the posteriors for the target domain in an unsupervised way. This is particularly valuable when the target domain can change on the fly. Interestingly, this flexibility is achieved by keeping the annotated data collected in different source domains separate and by sharing only unlabeled data.
Finally, we test our unsupervised domain adaptation algorithm on a semantic segmentation task in real-world surveillance, and show its superiority compared to common heuristics.

\ifwacvfinal

\section*{Acknowledgements}

This work has been made possible thanks to the \emph{TRAIL} initiative (\url{https://trail.ac}). Part of it was supported by the Walloon region (Service Public de Wallonie Recherche, Belgium) under grant n°2010235 – \emph{ARIAC by DIGITALWALLONIA4.AI}. Anthony Cioppa is funded by the FNRS (\url{https://www.frs-fnrs.be/en/}).

\fi

\cleardoublepage 
{\small 

}

\cleardoublepage 
\appendix
\section{Supplementary material}\label{sec:supplementary}

\subsection{Priors of the source domains}

\begin{figure}
\begin{centering}
\includegraphics[width=\linewidth]{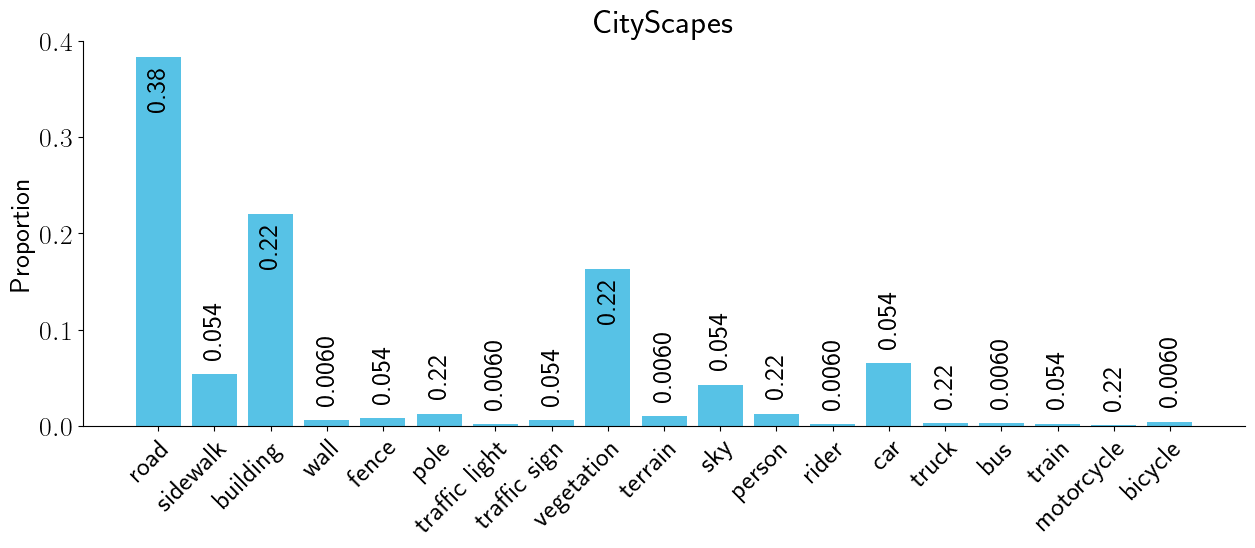}\,\\
\includegraphics[width=\linewidth]{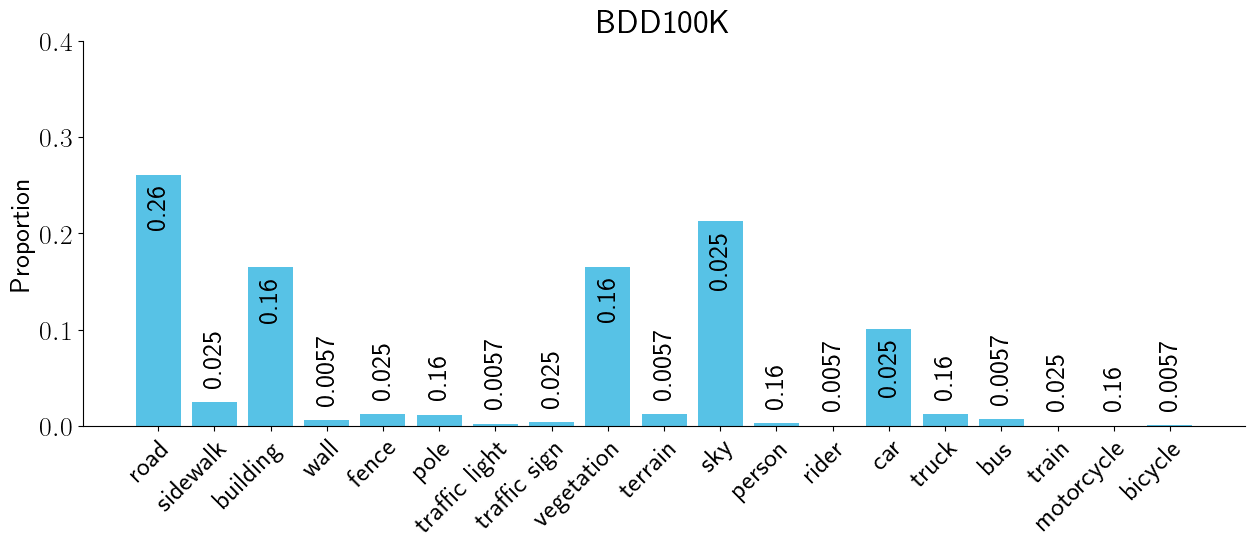}\,\\
\,\hfill{}\includegraphics[width=0.45\linewidth]{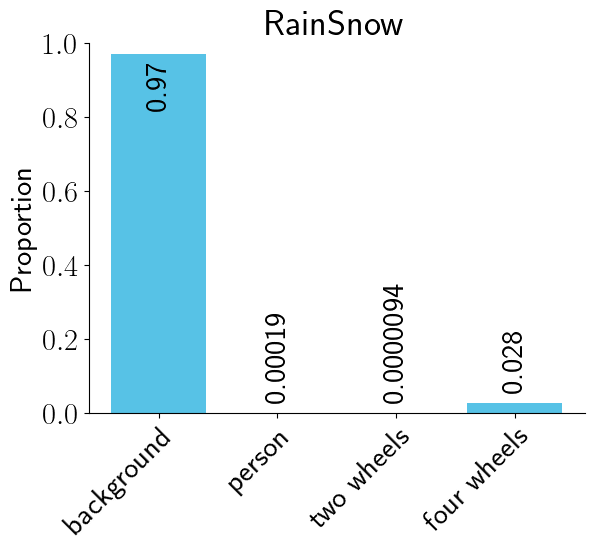}\hfill{}\includegraphics[width=0.45\linewidth]{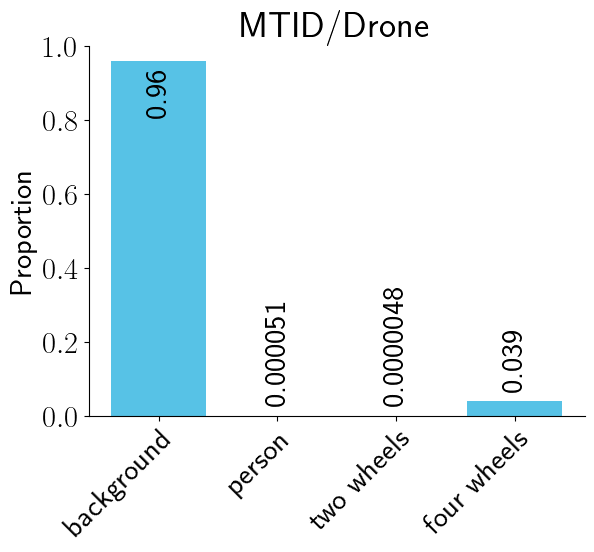}\hfill\,\\
\,\hfill{}\includegraphics[width=0.45\linewidth]{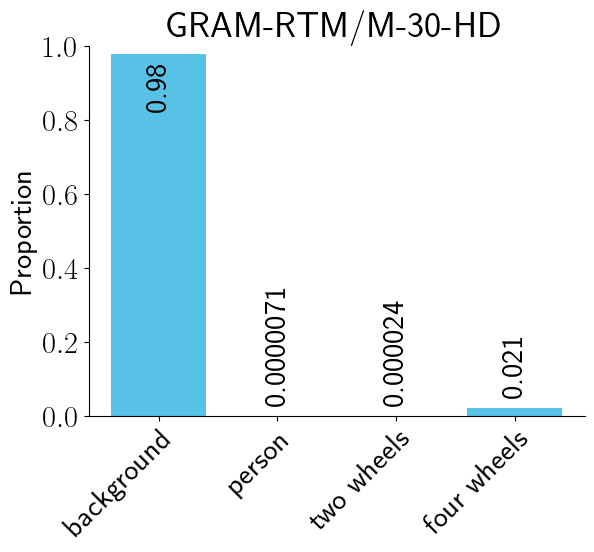}\hfill{}\includegraphics[width=0.45\linewidth]{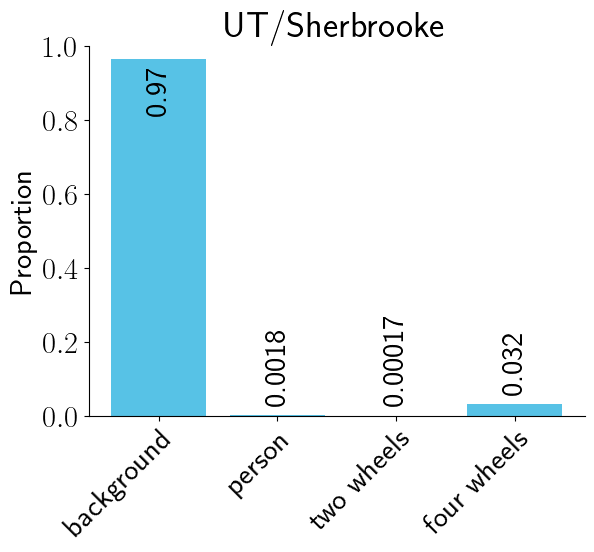}\hfill{}\,\\
\par\end{centering}
\caption{Histograms giving the priors, \ie the proportion of each semantic class, for each domain (from top to bottom and from left to right: CityScapes, BDD100K, RainSnow, MTID/Drone, GRAM-RTM/M-30-HD and UT/Sherbrooke). All numbers are given up to 2 significant figures.\label{fig:histograms}}
\end{figure}

In this section, we provide further details about the priors of each source domain used in our experiments, we provide the histograms with respect to the semantic classes for all domains (each corresponding to a dataset) in Figure~\ref{fig:histograms}.
As can be seen, there is a huge class imbalance with some of the classes. Some of them are especially underrepresented, like two wheels on the RainSnow, MTID/Drone, GRAM-RTM/M-30-HD and UT/Sherbrooke datasets.

\subsection{Details about the networks}
\begin{figure}
    \centering
    \includegraphics[width=\linewidth]{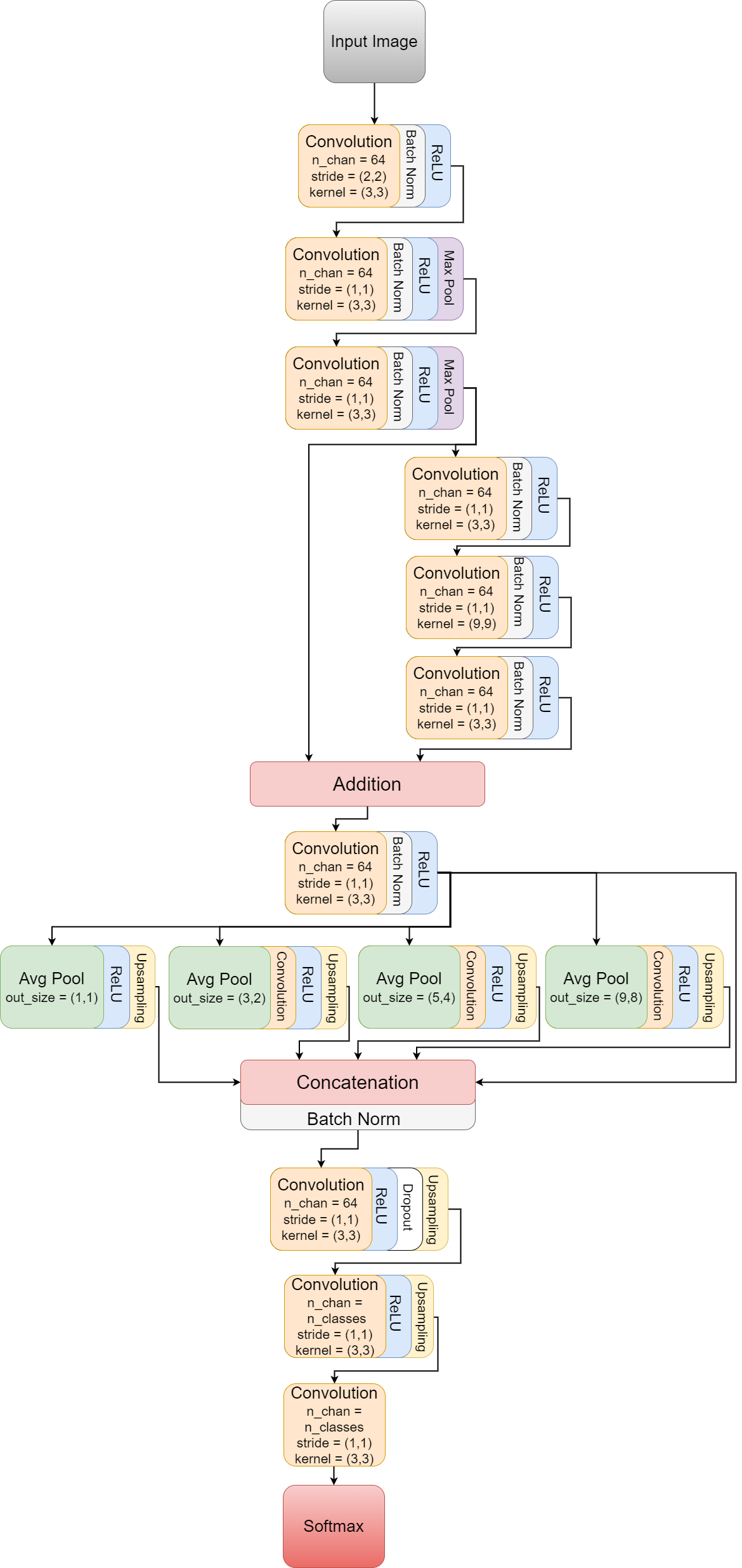}
    \caption{TinyNet architecture adapted to 720p images and an arbitrary number of output classes (either the semantic classes or the number of source domains).}
    \label{fig:tinynet}
\end{figure}

We use the TinyNet network for both the source and the domain discriminator models. This is a segmentation architecture first introduced in~\cite{Cioppa2018ABottomUp} for soccer player segmentation and later refined in~\cite{Cioppa2019ARTHuS}. Interestingly, due to its lightweight architecture, this network can be used for real-time inference on embedded devices, making it a great choice for real-world surveillance. Compared to the original architecture, we adapt it for the 720p input image size (compared to 1080p in the original publication). For that, only the values of the pooling layers applied in the pyramid pooling have to be adjusted. The architecture of the adapted network can be found in Figure~\ref{fig:tinynet}.

\subsection{Details about the domain discriminator model}
\begin{figure}
    \centering
    \includegraphics[width=\linewidth]{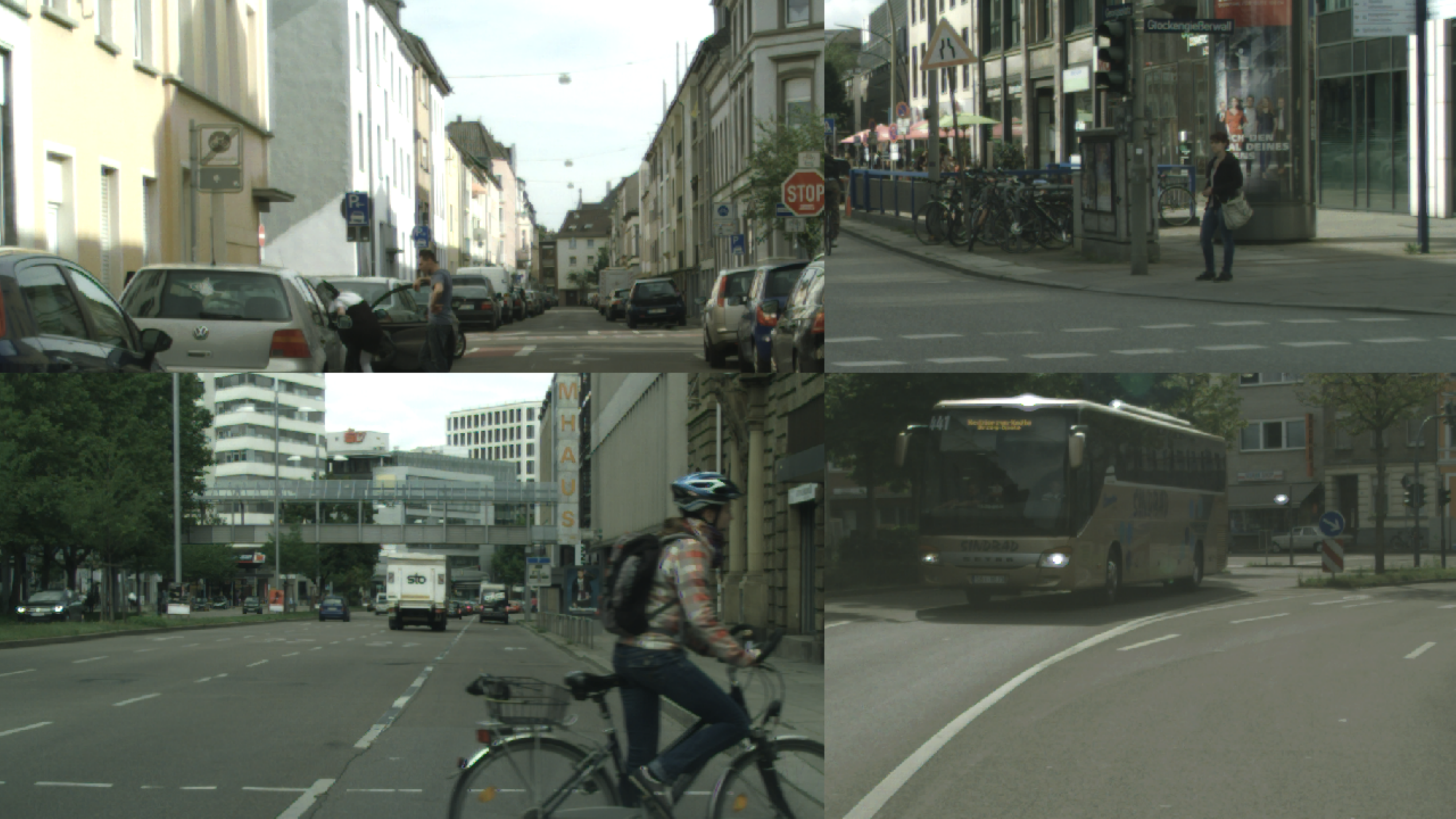}
    \caption{Example of multi-domain image used to train our domain discriminator model. The image is obtained with a mosaic transform for images of the CityScapes and BDD100K datasets.}
    \label{fig:mosaic_driving}
\end{figure}

Figure~\ref{fig:mosaic_driving} shows an example of multi-domain images that we generated to train our domain discriminator model. As can be seen, four patches cropped from four
randomly drawn images are combined to create training images, so that the domain discriminator model learns to recognise different domains within a same image. This is indeed an interesting feature since a target domain could combine some elements from different source domains (for example, the road of the target domain could look like the one from one source domain but the weather and lighting conditions may be those of another source domain).  

\subsection{Interpretation of posteriors}

The interpretation given to the posteriors depends on the priors. In particular, $\probabilityMeasure\left(\anHypothesis\mid\anEvidence\right)=\probabilityMeasure\left(\anHypothesis\right)$ when the evidence provides no information about the hypotheses. In fact, the priors can be seen as a reference, the posteriors being expressed \wrt it. 
Consider $\probabilityMeasure'$ and $\probabilityMeasure''$, two probability measures sharing common likelihoods $\probabilityMeasure'\left(\anEvidence\mid\anHypothesis\right)=\probabilityMeasure''\left(\anEvidence\mid\anHypothesis\right)$. The posteriors $\probabilityMeasure'\left(\anHypothesis\mid\anEvidence\right)$ and $\probabilityMeasure''\left(\anHypothesis\mid\anEvidence\right)$ convey the same information, but are expressed \wrt different references $\probabilityMeasure'\left(\anHypothesis\right)$ and $\probabilityMeasure''\left(\anHypothesis\right)$. The change of reference is termed \emph{prior shift} or \emph{target shift}. Assuming non-zero priors, it is given by~\cite{Sipka2022TheHitchhikerGuide,Pierard2014OnTheFly}
\begin{equation*}
    \probabilityMeasure''\left(\anHypothesis\mid\anEvidence\right)
    \propto
    \frac
    {\probabilityMeasure''\left(\anHypothesis\right)}
    {\probabilityMeasure'\left(\anHypothesis\right)}
    \,
    \probabilityMeasure'\left(\anHypothesis\mid\anEvidence\right)
\end{equation*}
and, when $\allHypotheses$ forms a partition of $\sampleSpace$, by
\begin{equation*}
    \probabilityMeasure''\left(\anHypothesis\mid\anEvidence\right)
    =
    \frac{
        \frac
        {\probabilityMeasure''\left(\anHypothesis\right)}
        {\probabilityMeasure'\left(\anHypothesis\right)}
        \,
        \probabilityMeasure'\left(\anHypothesis\mid\anEvidence\right)
    }{
        \sum_{\anHypothesis\in\allHypotheses}
        \frac
        {\probabilityMeasure''\left(\anHypothesis\right)}
        {\probabilityMeasure'\left(\anHypothesis\right)}
        \,
        \probabilityMeasure'\left(\anHypothesis\mid\anEvidence\right)
    }
    \dot
\end{equation*}

\subsection{Validation of predicted posteriors}

\begin{figure}
\begin{centering}
\includegraphics[width=\linewidth]{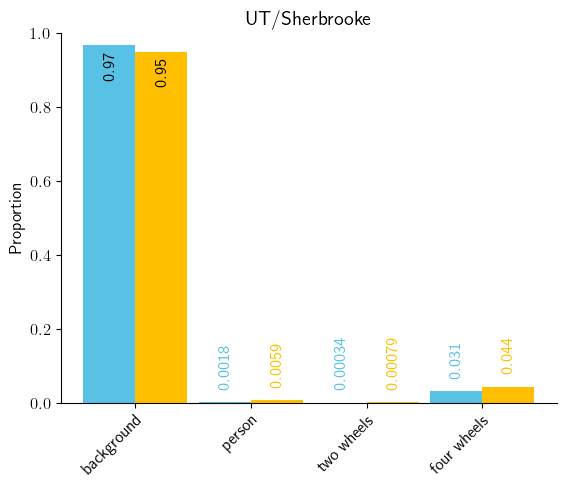}\,\
\par\end{centering}
\caption{Histogram for the UT/Sherbrooke dataset comparing the \mybluebis{priors of the test set (in blue)} and the \myorangebis{average of the posteriors obtained with our algorithm for each class (in orange)}. All numbers are given to 2 significant figures.\label{fig:histograms-posteriors}}
\end{figure}

\begin{figure}
\begin{centering}
\includegraphics[width=1.0\linewidth]{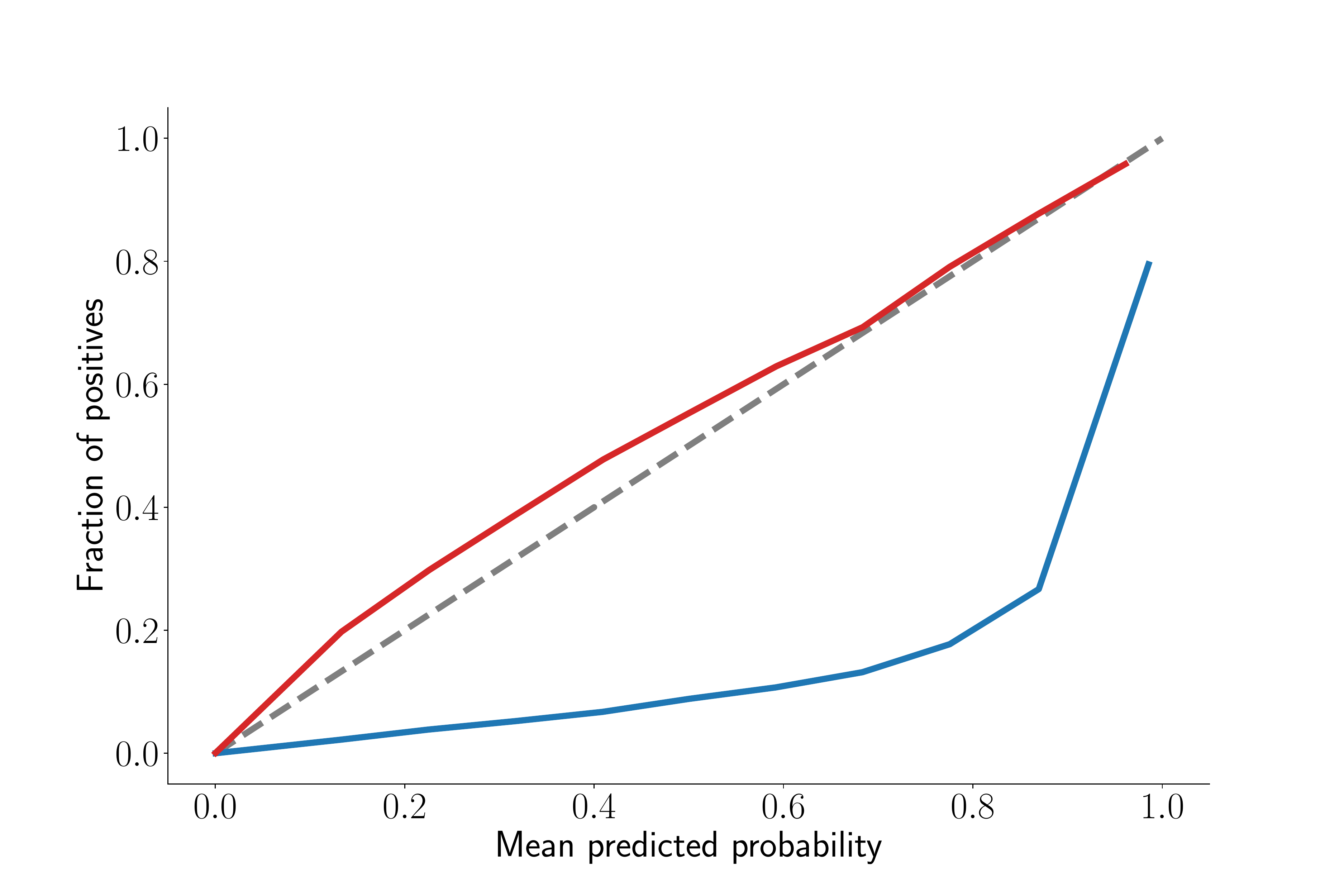}
\end{centering}
\caption{Calibration plot for the UT/Sherbrooke source model \myblue{before the target shift (in blue)} and \myred{after the target shift (in red)} for the "four wheels" class. The dotted line corresponds to the perfect case where the output of the model correspond to the posteriors. }\label{fig:calibration_plot_0}
\end{figure}

As stated in Section~\ref{subsec:ExperimentalSetup}, it is important that the source and domain discriminator models output trustworthy posteriors. To verify that, we perform two tests: (1)~we compare the class-by-class average of the posteriors estimated by our algorithm to the priors of the dataset—in particular, its test set (shown for one source model in Figure~\ref{fig:histograms-posteriors}), and (2)~we verify that our model is well calibrated using calibration plots (shown for one source model and one semantic class in Figure~\ref{fig:calibration_plot_0}). 

Since both the average of the posteriors is comparable to the priors of the dataset and the calibration plots are close to the $y=x$ line, we are confident that the outputs of our algorithm correspond to posteriors. We perform these tests for all source and discriminator models. 

\subsection{Extra results}

\begin{figure*}
\begin{centering}
\,\hfill{}\includegraphics[width=0.24\linewidth]{figures/figure_driving_accuracy.pdf}\hfill{}
\includegraphics[width=0.24\linewidth]{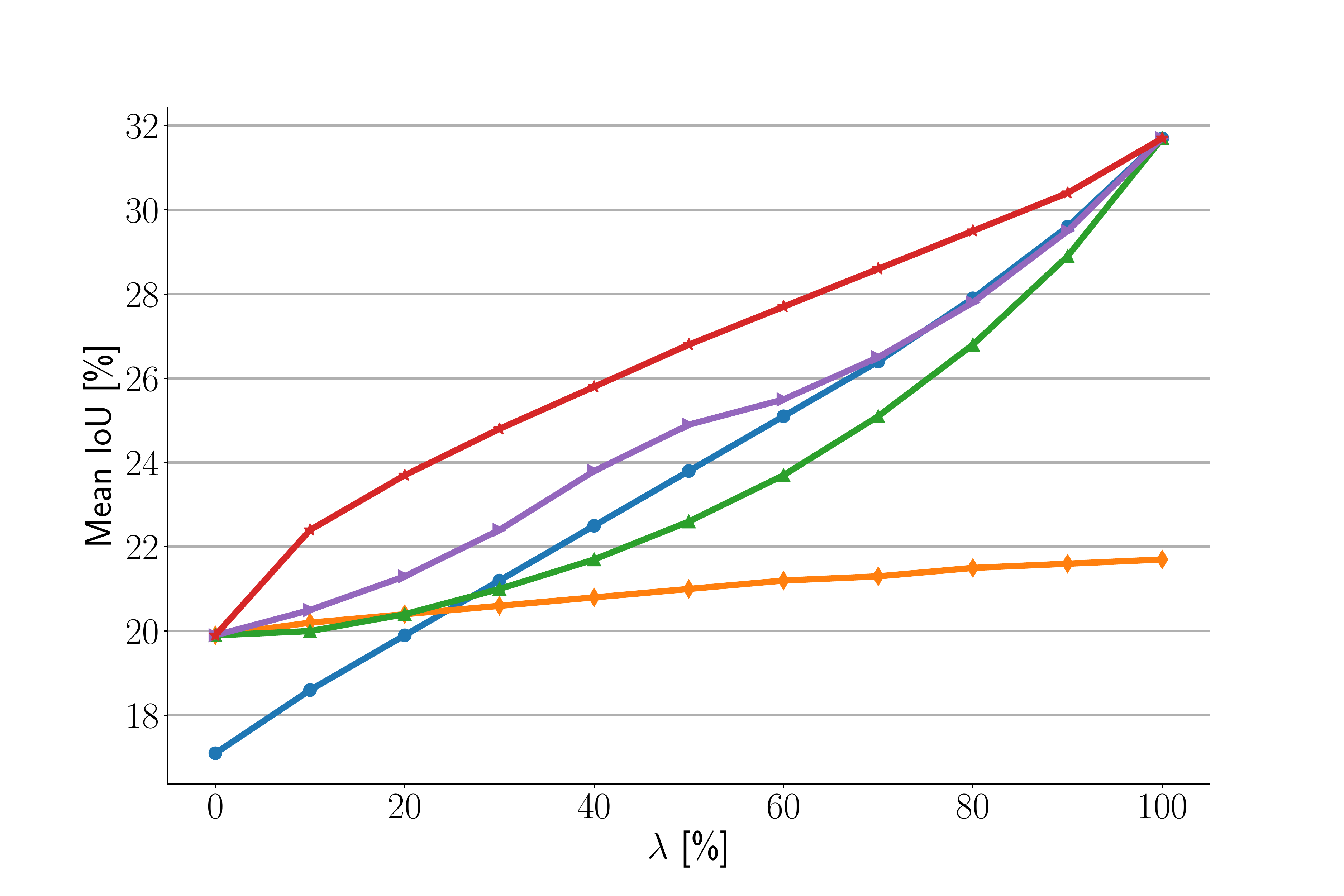}\hfill{}
\includegraphics[width=0.24\linewidth]{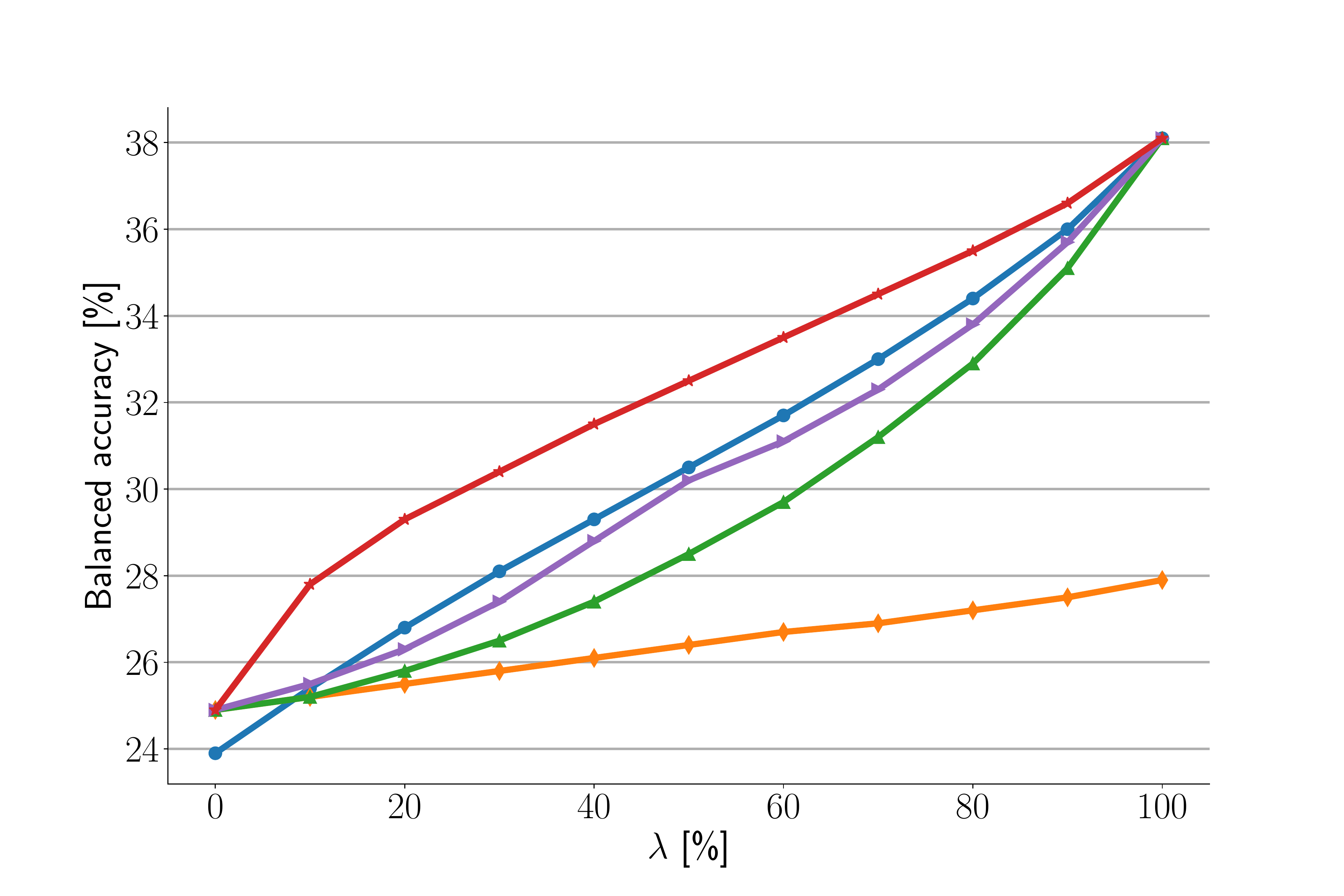}\hfill{}
\includegraphics[width=0.24\linewidth]{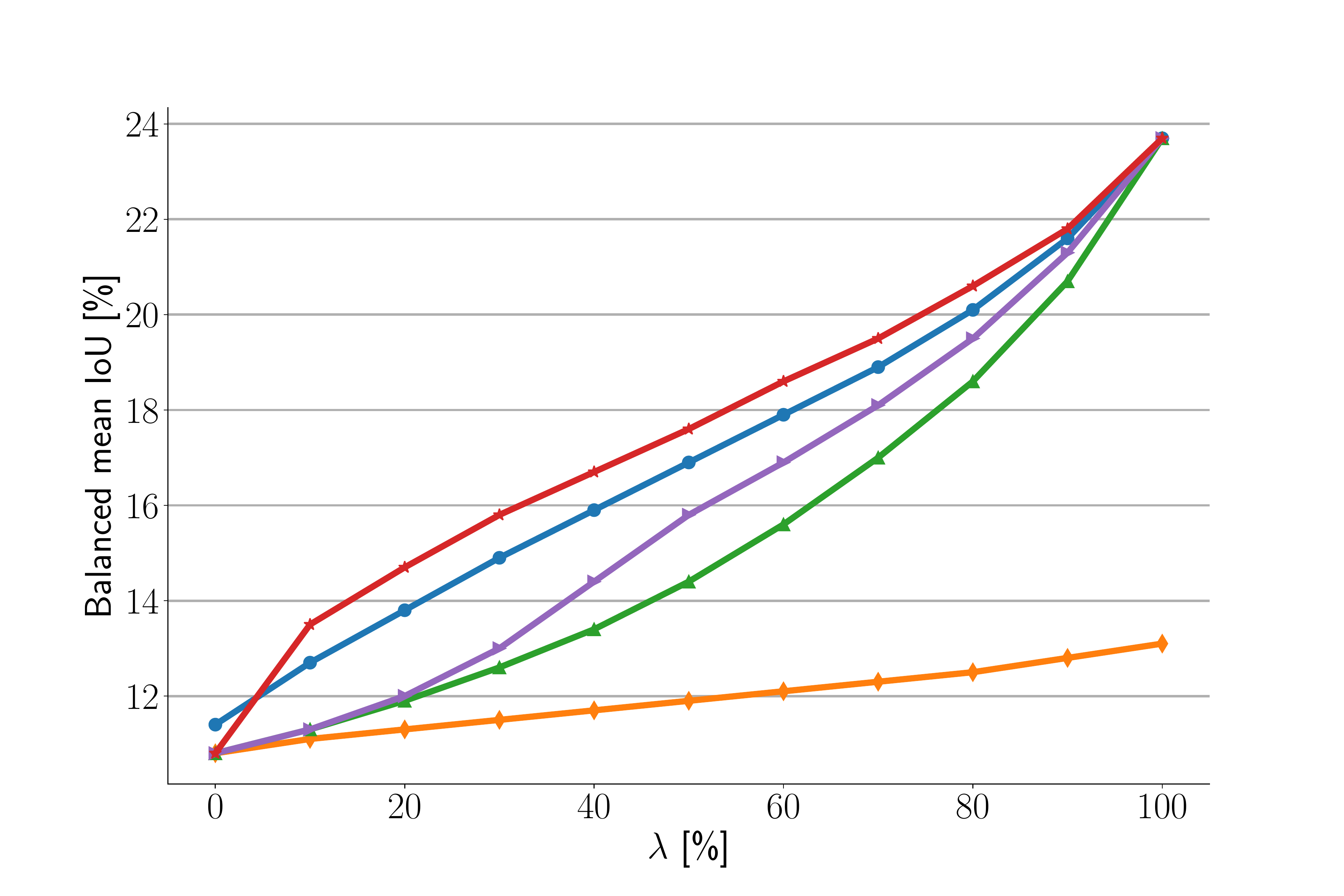}\hfill{}\,\\
\par\end{centering}
\caption{Results of our first experiment (all decisions made with the MAP strategy) showing from left to right: the accuracy, mean IoU, balanced accuracy, and balanced mean IoU. We compare the \myblue{source model trained on CityScapes (in blue)}, the \myorange{source model trained on BDD100K (in orange)}, the \mygreen{random selection of source models (in green)}, the \mypurple{linear combination of posteriors (in purple)}, and \myred{our algorithm (in red)}. $\weightUnconditional=0$ (resp. $=100)$ corresponds to BDD100K (resp. CityScapes) as target domain. \label{fig:driving_MAP}}
\end{figure*}

In this section, we provide extra results for both of our experiments. First, we provide the same graph as the one depicted in Figure~\ref{fig:driving_accuracy} for the $3$ remaining segmentation scores (\ie balanced accuracy, mean IoU, and balanced mean IoU) in Figure~\ref{fig:driving_MAP}. As can be seen, for all scores, our algorithm outperforms all other heuristics for each target domain. 

Next, we provide the performance of our algorithm for our second experiment with 4 source domains for different $\weightUnconditional_\idxDomainSource$ in Table~\ref{tab:results-exp4sources} for the accuracy and mean IOU, and in Table~\ref{tab:results-exp4sources-balanced} for the balanced accuracy and balanced mean IOU. We compare our algorithm with the four source models (one model for each source domain), the random selection of source models, and the linear combination of posteriors. Note that the histograms presented in Figure~\ref{fig:results-4sources} represent the column-by-column average values of these tables.

\begin{table*}[ht]
\renewcommand\cellalign{cc}
\scriptsize
    \caption{\textbf{Results of our second experiment (all decisions made with the MAP strategy) for various target domains obtained from 4 source domains for various $\weightUnconditional_\idxDomainSource$ for the accuracy/mean IoU scores. We compare the different source models (SM) trained on their source domain, the random selection of source models, the linear combination of posteriors, and our algorithm.}}
    \centering
    \setlength{\tabcolsep}{2pt}
    \resizebox{\linewidth}{!}{
    \begin{tabular}{c|c|c|c||c|c|c|c|c|c|c}
    
    \multicolumn{4}{c||}{Parameters} & \multicolumn{7}{c}{Methods} \\
    
    \midrule
    
    $\weightUnconditional_1$ & $\weightUnconditional_2$ & $\weightUnconditional_3$ & $\weightUnconditional_4$ & \makecell{$\text{SM}_1$ \\ RainSnow} & \makecell{$\text{SM}_2$ \\MTID/Drone} & \makecell{$\text{SM}_3$ \\ GRAM-RTM/M-30-HD} & \makecell{$\text{SM}_4$ \\UT/Sherbrooke} & \makecell{Random selection \\of source models} & \makecell{Linear combination \\ of posteriors} & \textbf{Ours} \\ 
    
    \midrule \midrule
    
    1 & 0 & 0 & 0 & $97.9/35.1$ & $96.4/26.3$ & $96.0/27.6$ & $96.1/28.3$ & \textcolor{gray}{$97.9/35.1$} & \textcolor{gray}{$97.9/35.1$} & \textcolor{gray}{$97.9/35.1$} \\
    \midrule
    0 & 1 & 0 & 0 & $96.8/33.3$ & $98.2/40.4$ & $95.5/31.3$ & $95.0/29.4$ & \textcolor{gray}{$98.2/40.4$} & \textcolor{gray}{$98.2/40.4$} & \textcolor{gray}{$98.2/40.4$} \\
    \midrule
    0 & 0 & 1 & 0 & $98.2/34.6$ & $97.8/31.8$ & $99.5/55.7$ & $96.1/28.2$ & \textcolor{gray}{$99.5/55.7$} & \textcolor{gray}{$99.5/55.7$} & \textcolor{gray}{$99.5/55.7$} \\
    \midrule
    0 & 0 & 0 & 1 & $96.2/32.9$ & $95.7/27.4$ & $96.5/32.3$ & $99.0/67.1$ & \textcolor{gray}{$99.0/67.1$} & \textcolor{gray}{$99.0/67.1$} & \textcolor{gray}{$99.0/67.1$} \\
    \midrule
    \sfrac{1}{2} & \sfrac{1}{2} & 0 & 0 & $97.4/\textbf{35.0}$ & $97.2/33.7$ & $95.7/29.6$ & $95.5/28.9$ & $97.3/34.0$ & $\textbf{97.7}/34.1$ & $97.0/31.7$ \\
    \midrule
    \sfrac{1}{2} & 0 & \sfrac{1}{2} & 0 & $98.0/35.3$ & $97.2/28.8$ & $97.8/38.0$ & $96.1/28.4$ & $97.9/36.9$ & $\textbf{98.5}/36.4$ & $98.0/\textbf{39.3}$ \\
    \midrule
    \sfrac{1}{2} & 0 & 0 & \sfrac{1}{2} & $97.1/33.5$ & $96.0/27.0$ & $96.2/30.0$ & $97.5/52.6$ & $97.3/42.5$ & $\textbf{98.2}/42.8$ & $\textbf{98.2}/\textbf{57.4}$ \\
    \midrule
    0 & \sfrac{1}{2} & \sfrac{1}{2} & 0 & $97.4/33.7$ & $98.0/36.8$ & $97.6/\textbf{43.1}$ & $95.6/28.9$ & $97.8/39.7$ & $\textbf{98.4}/39.2$ & $97.4/33.7$ \\
    \midrule
    0 & \sfrac{1}{2} & 0 & \sfrac{1}{2} & $96.5/33.6$ & $96.9/34.1$ & $96.0/31.9$ & $97.1/54.6$ & $97.0/43.0$ & $97.8/42.8$ & $\textbf{98.0}/\textbf{60.4}$ \\
    \midrule
    0 & 0 & \sfrac{1}{2} & \sfrac{1}{2} & $97.2/33.7$ & $96.8/29.6$ & $98.1/38.3$ & $97.5/54.1$ & $97.8/46.8$ & $\textbf{98.7}/43.5$ & $98.5/\textbf{62.0}$ \\
    \midrule
    \sfrac{1}{3} & \sfrac{1}{3} & \sfrac{1}{3} & 0 & $97.6/34.9$ & $97.4/33.2$ & $97.1/\textbf{36.4}$ & $95.8/28.8$ & $97.3/34.8$ & $\textbf{98.0}/34.5$ & $97.1/31.3$ \\
    \midrule
    \sfrac{1}{3} & \sfrac{1}{3} & 0 & \sfrac{1}{3} & $97.0/33.8$ & $96.7/31.7$ & $96.0/30.5$ & $96.7/48.4$ & $96.8/37.3$ & $97.6/35.4$ & $\textbf{97.7}/\textbf{55.3}$ \\
    \midrule
    \sfrac{1}{3} & 0 & \sfrac{1}{3} & \sfrac{1}{3} & $97.4/33.8$ & $96.7/28.4$ & $97.4/34.4$ & $97.0/47.9$ & $97.3/38.7$ & $\textbf{98.2}/36.4$ & $98.0/\textbf{56.3}$ \\
    \midrule
    0 & \sfrac{1}{3} & \sfrac{1}{3} & \sfrac{1}{3} & $97.0/33.7$ & $97.2/33.5$ & $97.3/35.8$ & $96.7/49.5$ & $97.1/39.3$ & $97.7/35.5$ & $\textbf{97.8}/\textbf{57.3}$ \\
    \midrule
    \sfrac{1}{4} & \sfrac{1}{4} & \sfrac{1}{4} & \sfrac{1}{4} & $97.2/33.9$ & $97.0/31.7$ & $96.9/33.7$ & $96.5/45.4$ & $96.9/35.9$ & $\textbf{97.8}/34.1$ & $97.6/\textbf{53.9}$ \\
    
    \bottomrule
    
    \end{tabular}}
    \label{tab:results-exp4sources}
\end{table*}
\begin{table*}[ht]
\renewcommand\cellalign{cc}
\scriptsize
    \caption{\textbf{Results of our second experiment (all decisions made with the MAP strategy) for various target domains obtained from 4 source domains for various $\weightUnconditional_\idxDomainSource$ for the balanced accuracy/balanced mean IoU scores. We compare the different source models (SM) trained on their source domain, the random selection of source models, the linear combination of posteriors, and our algorithm.}}
    \centering
    \setlength{\tabcolsep}{2pt}
    \resizebox{\linewidth}{!}{
    \begin{tabular}{c|c|c|c||c|c|c|c|c|c|c}
    
    \multicolumn{4}{c||}{Parameters} & \multicolumn{7}{c}{Methods} \\
    
    \midrule
    
    $\weightUnconditional_1$ & $\weightUnconditional_2$ & $\weightUnconditional_3$ & $\weightUnconditional_4$ & \makecell{$\text{SM}_1$ \\ RainSnow} & \makecell{$\text{SM}_2$ \\MTID/Drone} & \makecell{$\text{SM}_3$ \\ GRAM-RTM/M-30-HD} & \makecell{$\text{SM}_4$ \\UT/Sherbrooke} & \makecell{Random selection \\of source models} & \makecell{Linear combination \\ of posteriors} & \textbf{Ours} \\ 
    
    \midrule \midrule
    
    1 & 0 & 0 & 0 & $38.7/21.0$ & $28.0/9.5$ & $30.8/12.2$ & $31.9/13.4$ & \textcolor{gray}{$38.7/21.0$} & \textcolor{gray}{$38.7/21.0$} & \textcolor{gray}{$38.7/21.0$} \\
    \midrule
    0 & 1 & 0 & 0 & $37.0/18.8$ & $46.7/29.6$ & $37.0/18.6$ & $34.0/15.6$ & \textcolor{gray}{$46.7/29.6$} & \textcolor{gray}{$46.7/29.6$} & \textcolor{gray}{$46.7/29.6$} \\
    \midrule
    0 & 0 & 1 & 0 & $39.1/18.8$ & $35.2/15.7$ & $58.8/39.0$ & $33.8/15.4$ & \textcolor{gray}{$58.8/39.0$} & \textcolor{gray}{$58.8/39.0$} &  \textcolor{gray}{$58.8/39.0$} \\
    \midrule
    0 & 0 & 0 & 1 & $38.7/19.7$ & $30.1/11.9$ & $36.7/17.1$ & $75.9/61.1$ & \textcolor{gray}{$75.9/61.1$} & \textcolor{gray}{$75.9/61.1$} & \textcolor{gray}{$75.9/61.1$} \\
    \midrule
    \sfrac{1}{2} & \sfrac{1}{2} & 0 & 0 & $\textbf{38.5}/\textbf{20.8}$ & $38.1/19.4$ & $34.2/15.5$ & $33.1/14.5$ & $38.3/20.1$ & $36.0/17.7$ & $34.2/15.9$ \\
    \midrule
    \sfrac{1}{2} & 0 & \sfrac{1}{2} & 0 & $39.2/20.4$ & $31.3/12.8$ & $43.1/25.2$ & $32.8/14.3$ & $41.1/22.8$ & $37.7/19.2$ & $\textbf{43.4}/\textbf{25.3}$ \\
    \midrule
    \sfrac{1}{2} & 0 & 0 & \sfrac{1}{2} & $38.0/19.0$ & $29.3/10.9$ & $33.9/14.7$ & $62.2/45.4$ & $50.1/32.0$ & $45.1/27.5$ & $\textbf{66.5}/\textbf{50.4}$ \\
    \midrule
    0 & \sfrac{1}{2} & \sfrac{1}{2} & 0 & $37.6/18.0$ & $41.7/22.2$ & $\textbf{48.3}/\textbf{30.3}$ & $33.7/15.1$ & $45.0/26.2$ & $41.3/22.6$ & $37.6/18.3$ \\
    \midrule
    0 & \sfrac{1}{2} & 0 & \sfrac{1}{2} & $38.5/19.3$ & $38.6/20.1$ & $36.9/17.2$ & $63.4/46.7$ & $51.0/33.0$ & $45.3/27.8$ & $\textbf{70.1}/\textbf{54.5}$ \\
    \midrule
    0 & 0 & \sfrac{1}{2} & \sfrac{1}{2} & $39.0/19.8$ & $32.3/13.9$ & $42.4/22.7$ & $64.6/48.2$ & $53.5/35.2$ & $45.0/27.1$ & $\textbf{74.0}/\textbf{58.8}$ \\
    \midrule
    \sfrac{1}{3} & \sfrac{1}{3} & \sfrac{1}{3} & 0 & $38.5/19.7$ & $37.2/18.2$ & $\textbf{41.7}/\textbf{23.6}$ & $33.2/14.7$ & $39.1/20.5$ & $36.2/17.5$ & $34.3/15.6$ \\
    \midrule
    \sfrac{1}{3} & \sfrac{1}{3} & 0 & \sfrac{1}{3} & $38.1/19.0$ & $35.5/17.0$ & $35.1/15.7$ & $56.5/39.1$ & $43.3/24.9$ & $37.4/19.1$ & $\textbf{63.6}/\textbf{47.3}$ \\
    \midrule
    \sfrac{1}{3} & 0 & \sfrac{1}{3} & \sfrac{1}{3} & $38.3/19.2$ & $30.9/12.5$ & $38.8/19.4$ & $56.8/39.7$ & $44.6/26.0$ & $37.9/19.5$ & $\textbf{66.1}/\textbf{49.9}$ \\
    \midrule
    0 & \sfrac{1}{3} & \sfrac{1}{3} & \sfrac{1}{3} & $38.4/19.1$ & $37.6/19.1$ & $40.6/20.9$ & $57.8/40.7$ & $45.3/26.7$ & $38.2/19.9$ & $\textbf{66.3}/\textbf{49.9}$ \\
    \midrule
    \sfrac{1}{4} & \sfrac{1}{4} & \sfrac{1}{4} & \sfrac{1}{4} & $38.1/18.9$ & $35.3/16.8$ & $38.4/19.0$ & $53.0/35.5$ & $41.2/22.5$ & $35.8/17.4$ & $\textbf{62.0}/\textbf{45.1}$ \\

    \bottomrule
    
    \end{tabular}}
    \label{tab:results-exp4sources-balanced}
\end{table*}

\end{document}